%% file: sample_paper.tex
\documentclass[twoside]{article}

\usepackage[accepted]{aistats2023}

\usepackage[round]{natbib}
\renewcommand{\bibname}{References}
\renewcommand{\bibsection}{\subsubsection*{\bibname}}

\usepackage{tablefootnote}
\usepackage{tabularx}
\usepackage[dvipsnames]{xcolor}

\usepackage{microtype}
\usepackage{graphicx}
\usepackage{booktabs} 
\usepackage{algorithm}
\usepackage[noend]{algpseudocode}
\usepackage{amsfonts}
\usepackage{amsmath}
\usepackage{amsthm}
\usepackage{multirow}
\usepackage{bbding}
\usepackage{bbold}
\usepackage{wasysym}
\usepackage{amssymb}
\usepackage{pifont}
\usepackage[font=small]{caption}
\usepackage{hyperref}
\usepackage{isomath}
\usepackage{mathtools}
\usepackage{setspace}
\usepackage{thm-restate}
\usepackage{url}
\usepackage{wrapfig}
\input{math_commands.tex}

\usepackage{dsfont}
\usepackage{hyperref}
\usepackage{url}
\usepackage{enumitem}
\usepackage{amsfonts}

\usepackage{caption}
\usepackage{subcaption}

\theoremstyle{definition}
\newtheorem{definition}{Definition}[section]

\theoremstyle{property}

\theoremstyle{assumption}

\begin{document}

%

%

\twocolumn[

\aistatstitle{Improving Adaptive Conformal Prediction Using Self-Supervised Learning}

\runningauthor{Nabeel Seedat*, Alan Jeffares*, Fergus Imrie, Mihaela van der Schaar}

\aistatsauthor{ Nabeel Seedat* \And Alan Jeffares*}
\aistatsaddress{ University of Cambridge \And  University of Cambridge}
\aistatsauthor{ Fergus Imrie \And Mihaela van der Schaar}
\aistatsaddress{University of California, Los Angeles \And University of Cambridge \\ The Alan Turing Institute} ]

\begin{abstract}
Conformal prediction is a powerful distribution-free tool for uncertainty quantification, establishing valid prediction intervals with finite-sample guarantees. 
To produce valid intervals which are also adaptive to the difficulty of each instance, a common approach is to compute normalized nonconformity scores on a separate calibration set. 
Self-supervised learning has been effectively utilized in many domains to learn general representations for downstream predictors. However, the use of self-supervision beyond model pretraining and representation learning has been largely unexplored.
In this work, we investigate how self-supervised pretext tasks can improve the quality of the conformal regressors, specifically by improving the adaptability of conformal intervals. 
We train an auxiliary model with a self-supervised pretext task on top of an existing predictive model and use the self-supervised error as an additional feature to estimate nonconformity scores. 
We empirically demonstrate the benefit of the additional information using both synthetic and real data on the efficiency (width), deficit, and excess of conformal prediction intervals.
\end{abstract}

\section{INTRODUCTION}

Much of machine learning (ML) research can be summarized as the task of minimizing a model's predictive error on unseen data. In the regression setting, this might correspond to minimizing the mean squared error. However, for real-world applications, particularly in high-stakes domains such as healthcare and finance, we seek various forms of model trustworthiness. One such factor considered in this work is estimates of, or even better guarantees, concerning a model's predictive uncertainty, i.e. quantifying the uncertainty in a model's prediction.

\begin{figure*}[t]

\includegraphics[width=0.99\textwidth]{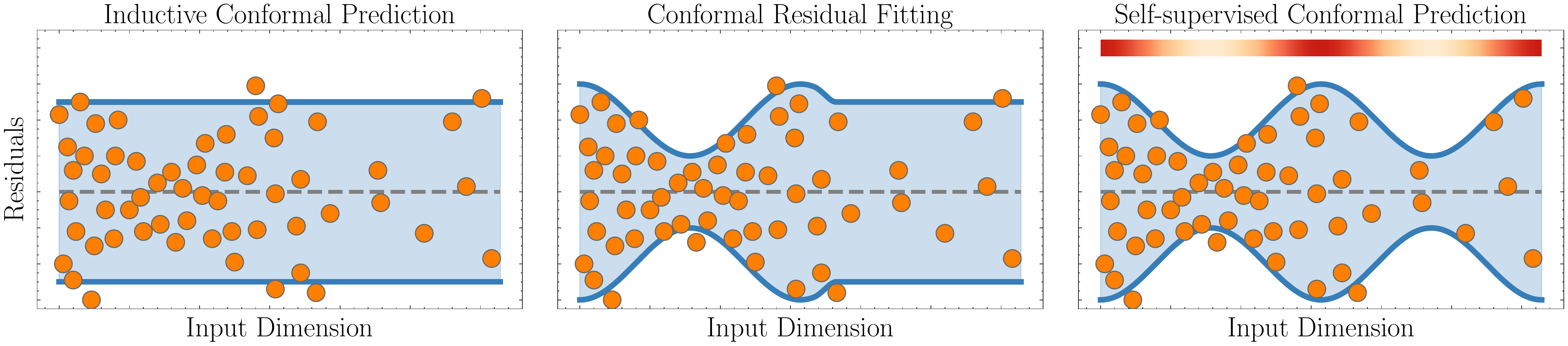}  
\caption{\textbf{Left.} Standard inductive conformal prediction results in constant width confidence intervals. \textbf{Center.} Conformal residual fitting produces adaptive intervals but can be inefficient in regions. \textbf{Right.} The errors of a self-supervised task are included above the plot with red indicating larger self-supervised errors. SSCP leverages these errors to improve the efficiency of conformal residual fitting. } \label{fig:intro_fig}
\end{figure*}

Conformal prediction \citep{vovk2005algorithmic} has received significant attention in recent years for this purpose. This powerful method provides valid prediction intervals with finite-sample, frequentist guarantees on the marginal coverage of the intervals. More precisely, for a random vector and label $(X, Y) \in \mathbb{R}^p \times \mathbb{R}$ drawn from a joint distribution $P_{XY}$ and a conformal predictor $\hat{C}: \mathbb{R}^p \to \{ \text{subsets in } \mathbb{R}\}$, then for a user-defined level of coverage ($1 -\alpha$), conformal prediction provides guarantees of the form
\begin{equation} \label{eqn:conformal_prediction}
    \mathbb{P}\{ Y \in \hat{C}(X) \} \geq 1 - \alpha .
\end{equation}
This method requires minimal assumptions on the data (i.e. exchangeability) and may be applied to any model upon which we can define a \textit{non-conformity score} quantifying the disagreement between the model output and ground truth target (e.g. mean squared error). It can even be applied post hoc to already trained models, assuming exchangeability of a calibration dataset with the test data of interest. For more details, see \citet{shafer2008tutorial}.

A widely used paradigm is inductive conformal prediction (ICP) \citealp{papadopoulos2008inductive, lei2018distribution}, which formulates conformal prediction using a dataset to train the predictive model and a calibration dataset on which to compute a critical non-conformity score, used to define the width of the prediction interval. However, despite guarantees of marginal coverage, these intervals are fixed and do not adapt to sample difficulty. In practice, we often desire wider intervals for challenging samples and narrower intervals for easier samples. Locally adaptive conformal prediction \citep{papadopoulos2008normalized,Johansson, lei2018distribution} aims to address this with a conformal normalization model, which predicts the residual errors of the predictive model (also known as conformalized residual fitting - CRF). The predicted residuals (representing sample prediction difficulty) are then used to adapt the intervals.

In this work, we tackle the case of ICP with locally adaptive intervals and seek to improve the quality of the prediction intervals. Although we primarily focus on the CRF approach, the core ideas of this work can be trivially adapted to other methods for local adaptivity such as Conformalized Quantile Regression (CQR) as we discuss in Sec. \ref{sec:CQR} \citep{romano2019conformalized}. However, a key benefit of the CRF approach is that the method does not directly modify the underlying base predictive model when adapting the prediction intervals. This enables post hoc conformalization in cases where we cannot augment or retrain such as a fixed/pre-trained regressor (e.g. behind an API) or in the case of a theoretically derived model (e.g. quantifying model error around the Navier-Stokes equations in fluid mechanics).

Specifically, we consider how to improve the conformal normalization function in order to improve the prediction intervals. To do so, we  diverge from the typical conformal prediction setting and consider how auxiliary self-supervised tasks can be leveraged to improve conformal predictors. Self-supervised learning has been employed effectively to learn representations to improve performance for downstream tasks; however, we also diverge from this standard setting of model pretraining and representation learning and address an under-explored avenue of using the error from a self-supervised task to help determine more challenging regions in input space such that conformal intervals can adapt appropriately.

Consequently, we propose \emph{Self-Supervised Conformal Prediction} (\textbf{SSCP}), a framework that provides a recipe to leverage information from self-supervised pretext tasks to improve prediction intervals (see Fig.~\ref{fig:intro_fig} for a contrast to alternative approaches). We note that the auxiliary self-supervision information does not impact the theoretical guarantees of conformal prediction (see Sec. \ref{guarantees_ss}).  To the best of our knowledge, this is the first work to integrate and examine the benefit of self-supervised errors to improve conformal prediction making the following contributions:

\textbf{Contributions:} \textbf{\textcolor{BrickRed}{\textcircled{1}}} \textbf{Self-supervision improves the residual estimates, especially in challenging and sparser regions:} We empirically demonstrate that self-supervised losses improve residual estimates thereby improving the prediction interval adaptiveness (Sec.\ref{synthetic}), especially for more challenging examples and in sparser regions (Sec.\ref{insights-exp}).
\textbf{\textcolor{BrickRed}{\textcircled{2}}} \textbf{Labeled data can be repurposed:} We empirically demonstrate that labeled data can be repurposed for the self-supervised task to improve the quality of conformal prediction intervals (Sec.\ref{label-exp}). 
\textbf{\textcolor{BrickRed}{\textcircled{3}}} \textbf{Unlabeled data improves CP:} We empirically demonstrate the value of unlabeled data to improve the quality of conformal prediction intervals beyond a standard application of self-supervision representation learning (Sec.\ref{unlabel-exp}).

\section{RELATED WORK} 
\paragraph{Adaptive Conformal Prediction.} While standard conformal prediction does provide marginal coverage (Equation \ref{eqn:conformal_prediction}), in practice we often desire conditional coverage $\mathbb{P}\{ Y \in \hat{C}(X)| X = x \} \geq 1 - \alpha$.
Although it has been shown that exact conditional coverage cannot be guaranteed in finite samples \citep{foygel2021limits, vovk2012conditional}, many works have improved the situation by relaxing the requirement of guarantees on coverage or by maintaining marginal coverage whilst seeking to improve adaptivity to a given test point. 

Locally adaptive conformal prediction \citep{papadopoulos2008normalized, papadopoulos2011regression,johansson2015} uses an independent model to induce adaptiveness, hence its mechanism is independent of the base predictive model. This flexibility means it can be applied to any architecture of predictive model or to an already trained model.  

In contrast, typically, other methods are not independent of the predictive model as they require specialized model architectures to obtain adaptive intervals, they require additional assumptions, and/or they sacrifice coverage guarantees. For example, Conformalized Quantile Regression (CQR) \citealp{romano2019conformalized} has a modified neural network that predicts the two quantiles. Parallel lines of research consider non-parametric density estimation-based approaches \citep{chernozhukov2021distributional, izbicki2019flexible} and conformal histogram regression \citep{sesia2021conformal} which discretely approximates the conditional density $f(y|x)$. 

Finally, conformal prediction splits the dataset into training and calibration datasets. While this may not be an issue with sufficient labeled data, in many settings we only have access to small/limited labeled datasets, which might significantly affect performance. In such settings, we often have access to large amounts of unlabeled data. To the best of our knowledge, SSCP is the first work to take advantage of such unlabeled data for conformal prediction in the context of self-supervised learning. We note that SSCP requires no additional splits in the labeled data over the underlying ICP method.

\paragraph{Self-Supervised Learning.} 
Self-supervised learning has been effectively utilized in many domains, from computer vision \citep{chen2020simple,grill2020bootstrap} to NLP \citep{kenton2019bert,brown2020language} to learn informative representations from raw input features, with the goal of improving the performance of downstream models. In the tabular domain, masking and reconstruction methods are often popular pretext tasks, using methods such as autoencoder reconstruction, VIME \citep{yoon2020vime}, where the task is to recover a mask vector in addition to the original sample from a corrupted version of the input, and other variants, e.g. \citep{lee2022self}. 

Although self-supervision has had great success, its use beyond model pretraining and representation learning has been largely unexplored. This work bridges this gap by studying how self-supervised losses can be used to provide additional information about the difficulty of a sample, making use of this information to improve the quality of conformal prediction intervals.

\section{BACKGROUND: Conformal Prediction}
In this section, we provide a brief summary of both ICP and CRF, two methods that are fundamental to our proposed framework. For proofs of the theoretical properties referenced in this section, we refer the reader to  \cite{vovk2012conditional}.

We begin with a brief description of the general setting. Consider the supervised learning setting in which we are provided with \textit{features} $\mathcal{X} \subseteq \mathbb{R}^{d_X}$ and \textit{labels} $\mathcal{Y} \subseteq \mathbb{R}^{d_Y}$. We wish to learn a prediction interval $\hat{C}: \mathcal{X} \to \{\text{subsets of } \mathbb{R}\}$ such that, for a desired coverage rate $1 - \alpha \in \mathbb{R}$ and a test point $(X, Y) \in \mathcal{X} \times \mathcal{Y}$, we ensure that Equation \ref{eqn:conformal_prediction} holds. To train such a model, we assume access to labeled data $\mathcal{D}_\text{labeled} \subset \mathcal{X} \times \mathcal{Y}$. We consider regression predictive tasks; hence, the label $y_i \in \mathcal{Y}$ is a scalar.

Throughout this work, we consider \textit{inductive} (or \textit{split}) conformal prediction. The standard practice in the literature is to subdivide the labeled data into training, calibration and testing $\mathcal{D}_\text{labeled} = \left\{\mathcal{D}_\text{train} \cup \mathcal{D}_\text{cal} \cup \mathcal{D}_\text{test} \right\}$. One might wish to also consider including an (optional) fourth subdivision $\mathcal{D}_\text{res}$, however, we defer considering this detail until Sec. \ref{sec:datasplits}. In this setting, the conformal predictor $f$, around whose predictions we desire confidence intervals, is trained on $\mathcal{D}_\text{train}$.

\subsection{Inductive conformal prediction (ICP)}
Given the trained predictive model $f$, we wish to produce conformal predictive intervals as a measure of uncertainty, $[l(x), r(x)]$ for each test instance $x \in \mathcal{D}_\text{test}$. These intervals should guarantee that the target $y$ of an instance $x$ lies within the interval such that $\mathbb{E}\left[1_{y \in\left[l(x), r(x) \right]}\right]\geq 1-\alpha$ where the significance level $\alpha \in (0,1)$ can be chosen\footnote{We select $\alpha=0.1$ in our experiments.} i.e. valid intervals with desired marginal coverage. In performing ICP, we only require the mild assumption of exchangeability between $\mathcal{D}_\text{cal}$ and $\mathcal{D}_\text{test}$ as defined in Def. \ref{def:exchange}.

\begin{definition}[Exchangeability] \label{def:exchange}
A dataset of $n$ observations is exchangeable if the data points do not follow any particular order, i.e., all $n$ permutations are equiprobable. Exchangeability is a weaker assumption than i.i.d.; indeed, i.i.d. observations satisfy exchangeability.
\label{exchangeability}
\end{definition}

To obtain these intervals, we carry out a calibration step. $\mathcal{D}_\text{cal}$ is used to compute a non-conformity score $\mu$, which estimates how different a new instance looks from other instances.  In practice, absolute error is a popular non-conformity score (i.e. $\mu(x) =  |y - f(x)|$). We then obtain an empirical distribution of non-conformity scores $\{ \mu(x) \mid x \in \mathcal{D}_\text{cal} \}$ over the calibration instances. 

The critical non-conformity score $\epsilon$, then corresponds to the $\lceil(|\mathcal{D}_\text{cal}| + 1)(1-\alpha)\rceil$-th smallest residual from the set $\{ \mu(x) \mid x \in \mathcal{D}_\text{cal} \}$. Finally, we use $\epsilon$ to construct the intervals $[l(x), r(x)] = [f(x)-\epsilon,  f(x)+\epsilon]$.

\subsection{Conformal residual fitting (CRF)} \label{sec:crf}
Unfortunately, the width of the ICP confidence intervals described in the previous section are constant for all instances. While they do obtain valid coverage, they do not reflect the difficulty of individual samples and are inefficient as a result (see Fig. \ref{fig:intro_fig}). To remedy this, CRF proposes to produce locally adaptive intervals based on an estimate of the predictive model's residuals. 

Specifically, to enable locally adaptive intervals (i.e. locally adaptive conformal prediction), a normalized non-conformity function $\gamma$ is used (Equation \ref{eqn:norm_conformity}). The numerator is computed as before based on $\mu$, however, the function $\sigma$ in the denominator permits us to normalize the non-conformity score per sample (i.e. as an estimate of difficulty), where $\sigma$ is a \emph{conformal normalization model}.

\begin{align}\label{eqn:norm_conformity}
    \gamma(x) \coloneqq \frac{|y - f(x)|}{\sigma(x)},
\end{align}

The critical non-conformity score $\epsilon$ is then computed similarly to before using the $\lceil(|\mathcal{D}_\text{cal}| + 1)(1-\alpha)\rceil$-th smallest residual of the empirical distribution of normalized non-conformity scores $\{\gamma(x) \mid x \in \mathcal{D}_\text{cal}\}$. 

To train $\sigma$, a sensible approach is to use an additional, typically disjoint dataset $\mathcal{D}_\text{res}$ upon which we can evaluate the trained predictive model $f$ to obtain predictions which will be used as targets. This allows us to construct the set of tuples $\{(x, |y - f(x)|) \mid x \in \mathcal{D}_\text{res} \}$. This set is used to train the \emph{conformal normalization model}, $\sigma : X \rightarrow \R^+$, to predict the residuals. $\sigma$ is then applied at testing time to estimate the predictive model's residuals on samples in  $X \in \mathcal{D}_\text{test}$.

\begin{figure*}[t]

    \captionsetup[subfigure]{labelformat=empty}
    \centering
    \begin{subfigure}[b]{0.385\textwidth}
    \includegraphics[width=\textwidth]{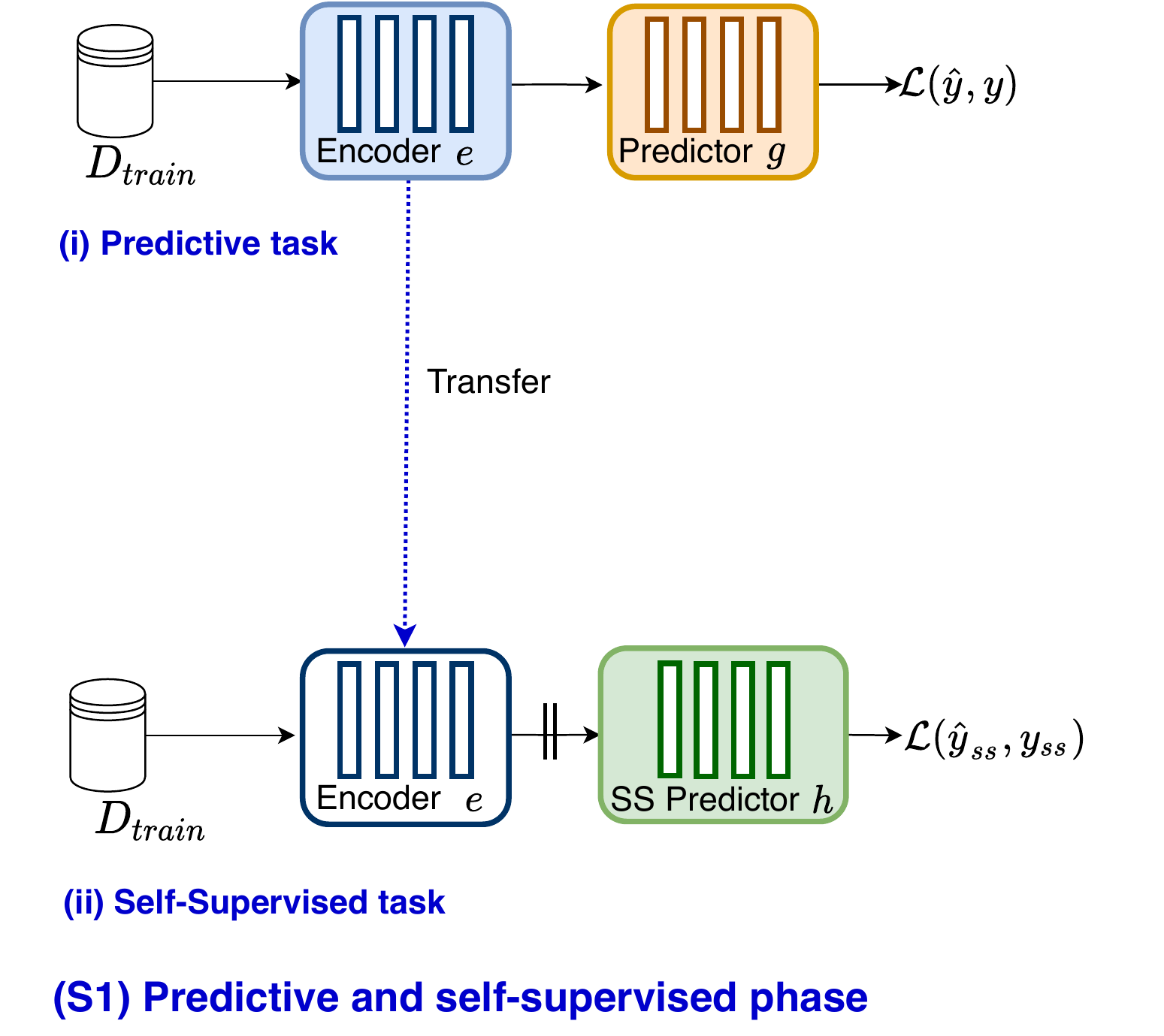}  
    \caption{}
    \end{subfigure}
    \hspace{-5mm}
    \begin{subfigure}[b]{0.385\textwidth}
    \includegraphics[width=\textwidth]{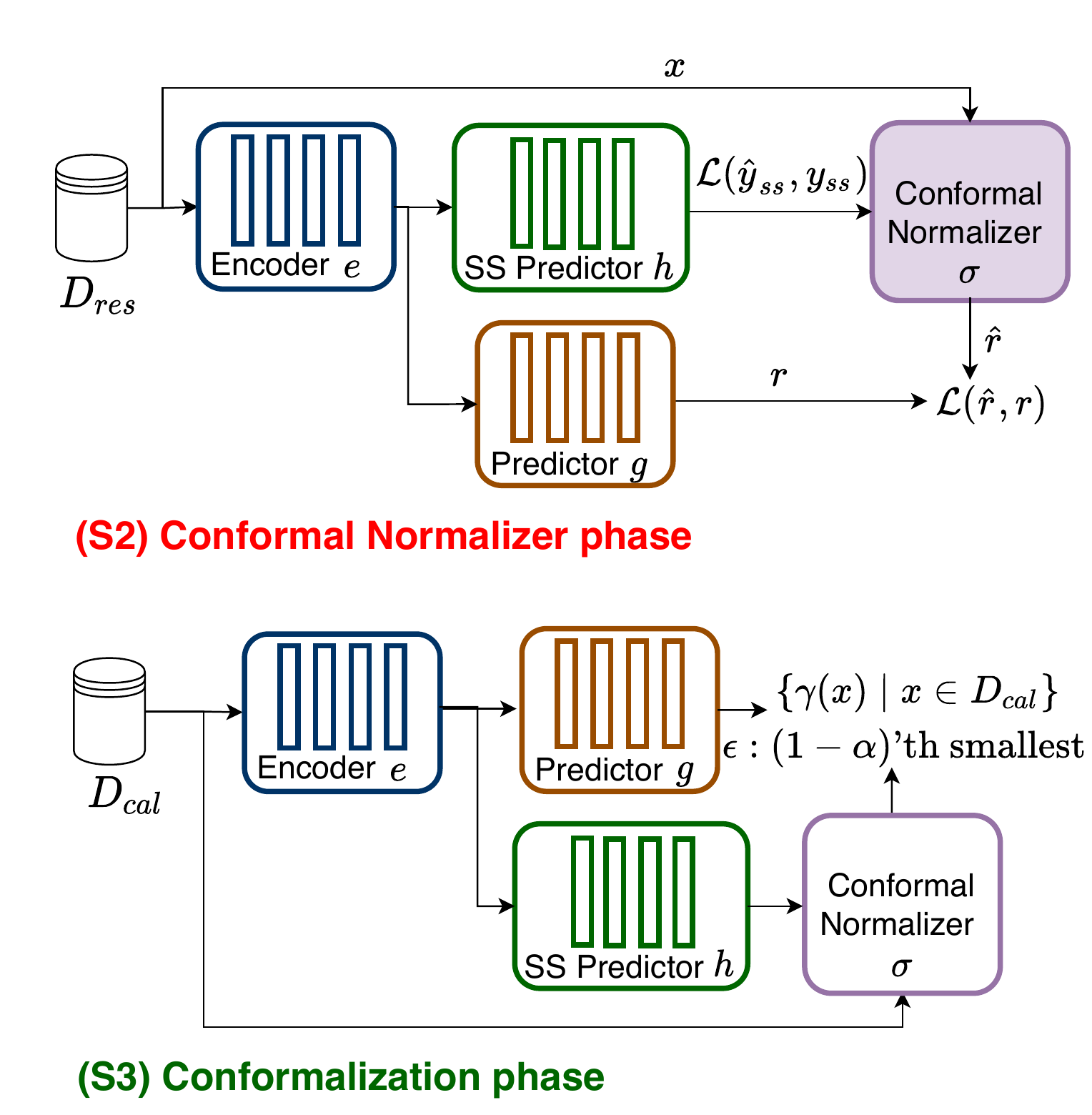}  
    \caption{}
    \end{subfigure}
    \hspace{-2mm}
    \begin{subfigure}[b]{0.25\textwidth}
    \includegraphics[width=\textwidth]{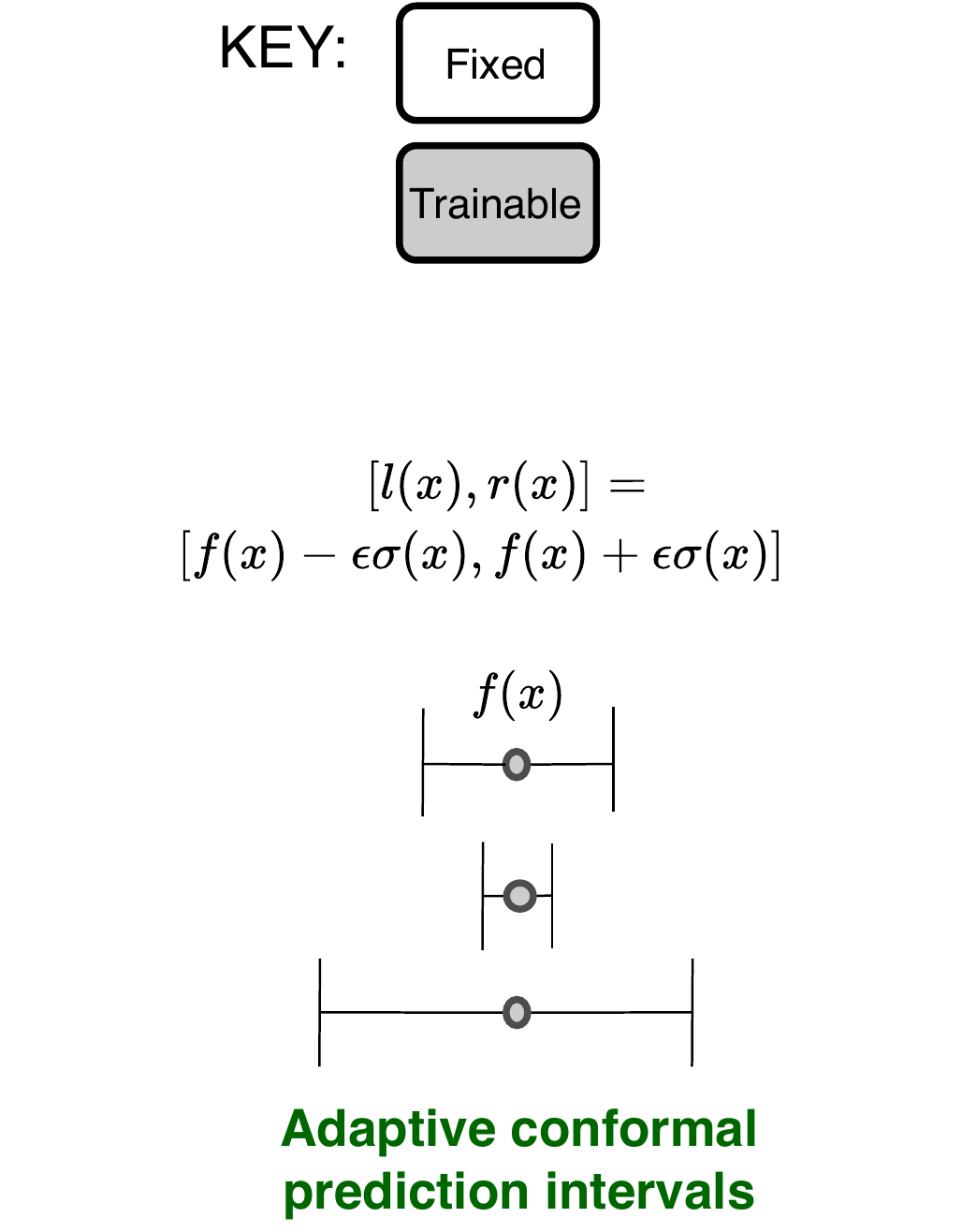}  
    \caption{}
    \end{subfigure}
    \caption{
  SSCP produces adaptive conformal prediction intervals leveraging self-supervised learning to improve the conformal normalizer ($\sigma$) and consists of 3 phases. \textcolor{blue}{(S1)} where \textcolor{blue}{(i)} train the predictive regressor $f=e \circ g $, thereafter \textcolor{blue}{(ii)} the encoder representation ($e$) is  transferred to use for the self-supervised task, where we train the self-supervised predictor $h$. \textcolor{red}{(S2)} the conformal normalizer phase is then carried out, with the conformal normalizer, augmented with the self-supervised input, trained to predict residuals (representing difficulty of prediction). Finally,  \textcolor{ForestGreen}{(S3)} the conformalization phase, computes non-conformity scores ($\mu$) on $\mathcal{D}_{cal}$, with the critical non-conformity score ($\epsilon$) chosen as the ($1-\alpha$)-smallest.  At test time, the adaptive prediction intervals [$l(x), r(x)$] are produced by adapting the critical non-conformity score $\epsilon$ using the conformal normalizer, i.e $\epsilon\sigma(x)$. The ``shaded blocks'' in each phase are trainable.
     }
\label{fig:idea}

\end{figure*}

\section{METHOD: Self-Supervised Conformal Prediction}\label{method}

In this section, we describe our proposed method to leverage self-supervision to improve locally adaptive conformal prediction, which we refer to as Self-Supervised Conformal Prediction (SSCP).  

Recall CRF adapts confidence intervals by predicting the residuals of the main model using an auxiliary residual model. By predicting regions in which the predictive model will make greater or smaller errors, the intervals can be adapted appropriately. However, the task of predicting residuals can be challenging; indeed, presumably they would not be residuals in the first place otherwise. Therefore, we propose to improve the performance of the residual model with the additional signal. 

Our method is motivated by the following idea:
can self-supervised tasks provide this added signal? With an appropriately selected self-supervised task, not only do we get its prediction at test time but also its ground truth target. Therefore, we have access to the self-supervised error \textit{even} at test time. If this error has a relationship with the error of the main model, then it can provide a useful input feature to the residual model. While correlated errors will often occur naturally due to factors such as noise or complexity in the data distribution, we can encourage this relationship in errors by careful selection of the self-supervised task and/or shared representation between the main model and the self-supervised model. 

Furthermore, the self-supervised model can simply be trained on the labeled training data as an extra step or, even better, it can also leverage any additional unlabeled data that is available. We highlight that this method can be applied orthogonally and \textit{in addition to} any standard self-supervised representation learning of the main model.

Our proposed SSCP method consists of three steps which we also summarize graphically in Fig.~\ref{fig:idea} and algorithmically in Algorithm \ref{algorithm_sscp}.
\begin{itemize}
    \item \emph{(S1) Predictive and self-supervised phase}: First, we train a predictive model to solve the regression task, as normal. Then, we share the learnt representations from the predictive model, and solve the self-supervised pretext task on top of these representations.
    \item \emph{(S2) Conformal normalizer phase}: Train a conformal normalizer to learn the residuals of the predictive model using the data features as well as the self-supervised errors on the same instances. 
    \item \emph{(S3) Conformalization phase}: Finally, we apply CRF using the prediction model, self-supervised model, and residual model trained in the previous steps.
\end{itemize}
In the following sections, we describe each step in more detail. First, however, we take a brief detour to describe the data splits and any assumptions on their distributions.

\subsection{Data splits and assumptions} \label{sec:datasplits}
We begin by briefly summarizing the required data splits for performing SSCP. We also highlight that our method requires no further data splits beyond CRF and is therefore \textit{more} data efficient due to its reuse of labeled data and ability to additionally use unlabeled data for the self-supervised tasks. $\mathcal{D}_\text{train}$ refers to the data on which the predictive and self-supervised models are trained. This may consist of a mixture of labeled and unlabeled data, where in many practical cases we have access to an unlabeled set with cardinality significantly greater than that of the labeled set. We also note that neither model is required to use this full set for training (e.g. the predictive model might only use the labeled data while the self-supervised model uses both labeled and unlabeled). 

Although not typically discussed in the literature, the residual model should ideally be trained on disjoint data from the predictive model. We refer to this as $\mathcal{D}_\text{res}$ and, by using a separate dataset, we ensure that the residual model does not learn on the potentially overfit $\mathcal{D}_\text{train}$ residuals.\footnote{We note that training both $f$ \& $\sigma$ on $\mathcal{D}_\text{train}$ is possible, but comes with the associated overfitting risk of training $\sigma$ on \textit{training} residuals instead of \textit{testing} residuals. We used a principled experimental design by including this additional split resulting in $\mathcal{D}_\text{res}$, but practitioners may obtain better performance by \textit{double dipping} on $\mathcal{D}_\text{train}$.}
Then, as usual, we have calibration and testing data $\mathcal{D}_\text{cal}$ and $\mathcal{D}_\text{test}$, respectively. The only assumption required on the data to achieve valid coverage is that these two datasets are exchangeable. We require no assumptions on the remaining data although, as in any learning problem, model performance is likely to be affected by a divergence between their distributions.

\subsection{Predictive and self-supervised phase (S1)} \label{sec:s1}

\paragraph{Predictive task.} 
The goal of the predictive stage is to obtain a regressor $f: \mathcal{X} \to \mathcal{Y} $ mapping inputs to targets. Typically, this is a machine learning model (e.g. a neural network) trained on $\mathcal{D}_\text{train}$. However, we place no assumptions on this model, which could be \textit{any} predictive model. An example of such a scenario beyond standard models might appear in a healthcare setting in which our model is the decision process of a doctor around which we would like to obtain valid confidence intervals. 

In this work, we focus on the more conventional setting of a data-based model trained on $\mathcal{D}_\text{train}$ to minimize the supervised loss $\mathcal{L}$ (usually mean squared error), i.e. $\min\limits_{f} \mathbb{E}_{(x,y) \sim \mathcal{D}_\text{train}}[\mathcal{L}(y,f({\mathbf{x}))] } $. We select this model to be a neural network consisting of an encoder $e$ and a predictor $g$ such that $f = g \circ e$. This provides us with the ability to share an encoding representation between the supervised and self-supervised task, which enforces a relationship between the two models. 

We also highlight that a standard application of self-supervised learning could be applied in learning the predictive model \textit{in addition to} the SSCP procedure we describe in this work. This would consist of applying standard self-supervision prior to the predictive task (S1). Our SSCP framework is orthogonal to such an application.

\paragraph{Self-supervised task.} The self-supervised task consists of training some model $f_{ss}$ also on $\mathcal{D}_\text{train}$ but this time for some pretext task. We also place no restrictions on the choice of self-supervised task. However, we wish its errors to have a useful relationship with those of the predictive model. 

In this work, we focus on tabular data where tasks such as input reconstruction-based tasks (such as autoencoders or VIME \citealp{yoon2020vime}) are popular. Depending on the task, various domain-specific tasks have been proposed in the self-supervised literature which may be leveraged in this step. We also use a neural network $f_{ss} = h \circ e$ where $e$ is the pre-trained encoder of the predictive model and $h$ is a task-specific decoder. We fix the weights of $e$, such that it retains the desired connection with the predictive task. We then optimize $h$ to solve the pretext task, minimizing the self-supervised loss $\mathcal{L}_{ss}$ as part of the following objective: 
\begin{align}\label{lss}
    \min\limits_{h} \mathbb{E}_{(X)\sim \mathcal{D}_\text{train}} [\mathcal{L}_{ss}(\mathbf{x},(h \circ e)(\mathbf{x}))].
\end{align}

\subsection{Conformal normalizer phase (S2)} \label{sec:s2}
The \emph{conformal normalization model} $\sigma$ is directly responsible for the adaptiveness of the prediction intervals. Hence, by improving $\sigma$, we can in turn improve the adaptiveness of the intervals. As previously discussed, this model is fit to $\mathcal{D}_\text{res}$ to avoid learning the training residuals, which may not generalize. In this work, we propose to incorporate the self-supervised error $\mathcal{L}_{\text{ss}}$ as an additional feature to $\sigma$. Fundamental to this, we can compute $\mathcal{L}_{\text{ss}}$ at both training and test time, since we do not need access to labels. 

Furthermore, to reiterate, if the self-supervised loss for a specific sample has a relationship to the difficulty of the predictive task, then, by adding $\mathcal{L}_{\text{ss}}$ as a feature, we pass this information to $\sigma$ when predicting the normalizing residuals. To do so, we augment the input space of $\sigma$ as $\hat{X} = \{X, \mathcal{L}_{\text{ss}}(X)\}$ and train $\sigma$ as usual. The decision of which residual normalizer model to use should be selected on a problem-specific basis. We successfully apply a random forest (Sec. \ref{synthetic}) and a neural network (Sec. \ref{real}) in our experiments.

\subsection{Conformalization phase (S3)}
Finally, given our trained predictive model $f$, self-supervised model $f_{ss}$, and residual model $\sigma$, we can apply the same calibration procedure as standard CRF on $\mathcal{D}_\text{cal}$ (as described in Sec.  \ref{sec:crf}) to obtain the $\alpha$-quantile non-conformity score $\epsilon$. Then for a new test example $x$, we obtain adaptive and valid intervals.
\begin{align}\label{instance_pred}
      [l(x), r(x)] = [f(x)-\epsilon  \sigma(\hat{x}), f(x) + \epsilon  \sigma(\hat{x})],
\end{align}
where $\hat{x} = \{x, \mathcal{L}_{\text{ss}}(x)\}$. We summarize the complete SSCP framework algorithmically in Algorithm \ref{algorithm_sscp}.

\vspace{1mm}
\begin{algorithm}

\caption{Self-supervised Conformal Prediction}
\begin{algorithmic}[1]
\State \textbf{Input:} $\mathcal{D}_\text{train}$, $\mathcal{D}_\text{res}$, $\mathcal{D}_\text{cal}$, $e$, $g$, $h$, $\sigma$
\Procedure{SSCP}{}      
    \State Train $f = g \circ e$ on  $\mathcal{D}_\text{train}$ (Sec.  \ref{sec:s1} S1 (i)) \;
    \State Train $f_{ss} = h \circ e$ on  $\mathcal{D}_\text{train}$ (Sec.  \ref{sec:s1} S1 (ii))\;
    \State Calculate self-supervised errors $\mathcal{L}_{\text{ss}}$ on $\mathcal{D}_\text{res}$ \;
    \State Train $\sigma$ on $\mathcal{D}_\text{res} \cup \mathcal{L}_{\text{ss}}$ (Sec.  \ref{sec:s2} S2)\;
    \State Apply standard CRF calibration (Sec.  \ref{sec:s2}) \;
    \State \textbf{Return:} $e$, $g$, $h$, $\sigma$ for test evaluation \;

\EndProcedure
\end{algorithmic}
\label{algorithm_sscp}
\end{algorithm}
\vspace{-1mm}

\subsection{Other locally adaptive methods} \label{sec:CQR}
The SSCP framework described throughout this section has primarily focused on CRF as a method for achieving locally adaptive intervals. However, simple adaptations to the procedure permit alternative methods to be considered. A powerful alternative that has emerged in recent years is CQR \citep{romano2019conformalized}. In Appendix \ref{app:CQR}, we describe the adaption required to integrate CQR into our SSCP framework and provide an empirical evaluation of CQR as an alternative method.

\subsection{Remark on SSCP and coverage guarantees} \label{guarantees_ss}
Conformal prediction provides theoretical guarantees on the validity of coverage of the prediction intervals based on the exchangeability assumption (see Def. \ref{def:exchange}).
SSCP inherits these same guarantees. Just as in the case of CRF, SSCP provides valid coverage provided the calibration set and test set satisfy exchangeability (see e.g. \cite{tibshirani2019conformal} for proof of these guarantees). We provide an extended comment on this point in Appendix \ref{app:coveragecomment}.

\section{EXPERIMENTS}\label{experiments}
In this section, we empirically assess the benefit provided by SSCP \footnote{https://github.com/seedatnabeel/SSCP}\footnote{https://github.com/vanderschaarlab/SSCP}  compared to CRF. 
We begin with a synthetic example to highlight \textit{why} adding a self-supervised loss to the conformal normalization model can improve prediction intervals.
Then, we demonstrate the performance of SSCP on a diverse range of real-world datasets, which differ in terms of numbers of samples, features, and difficulty.

\paragraph{Evaluation metrics}

We desire that the self-supervised information improves the quality of the conformal prediction intervals on average. We assess the quality based on the following commonly used metrics \citep{romano2019conformalized,navratil2020uncertainty}. Note, that in all cases, we still obtain intervals with valid coverage. Decreases in these metrics whilst still maintaining coverage implies superior adaptation and improvement to the quality of the prediction intervals:
\begin{itemize}
    \item Width: The size of the conformal prediction interval, i.e. $ \mathbb{E}[r(x) - l(x)], x \in X$
    \item Deficit: The interval shortfall, when the true value $y$ lies outside the predicted interval. i.e. $\mathbb{E}[1_{y \notin[l(x), r(x)]} \cdot \min \{|y-l(x)|,|y-r(x)|\}], x \in X$
    \item Excess: The additional width included not needed to capture the true value $y$. $ \mathbb{E}[1_{y \in [l(x), r(x)]} \cdot \min \{y-l(x), r(x)-y\}], x \in X$
\end{itemize}

\paragraph{Experimental setup.} We use the same underlying predictive model in all comparisons to isolate the effects of the conformal prediction method (since by definition a better predictive model can achieve narrower intervals). Our baseline is locally adaptive conformal prediction (denoted \textbf{CRF}), where the non-conformity score is the absolute error of the residuals.

In our experiments, we illustrate SSCP using VIME \citep{yoon2020vime} as an example of a potential self-supervision task. We selected VIME as it is regarded as highly performant on tabular data. As a pretext task, from a partially masked input, VIME aims to reconstruct the original sample in addition to recovering the mask vector. We provide additional experiments and ablations in Appendix \ref{appx-c}. Further experimental details can be found in Appendix \ref{appx-b}.

\subsection{Synthetic demonstration}\label{synthetic}        
We begin by illustrating our method on a synthetic dataset. While a standard application of self-supervision to improve the predictive model is likely to improve performance, the primary contribution of this work is to apply self-supervision to achieve a more adaptive conformal normalization model $\sigma$.

We isolate this effect by generating synthetic residuals from a hypothetical predictive model and comparing the performance of a conformal normalization model \emph{with} and \emph{without} access to the errors of a self-supervised task. Therefore, any improvements in this setting are exclusively due to this additional information. 

\begin{figure}[h]
\includegraphics[width=0.99\linewidth]{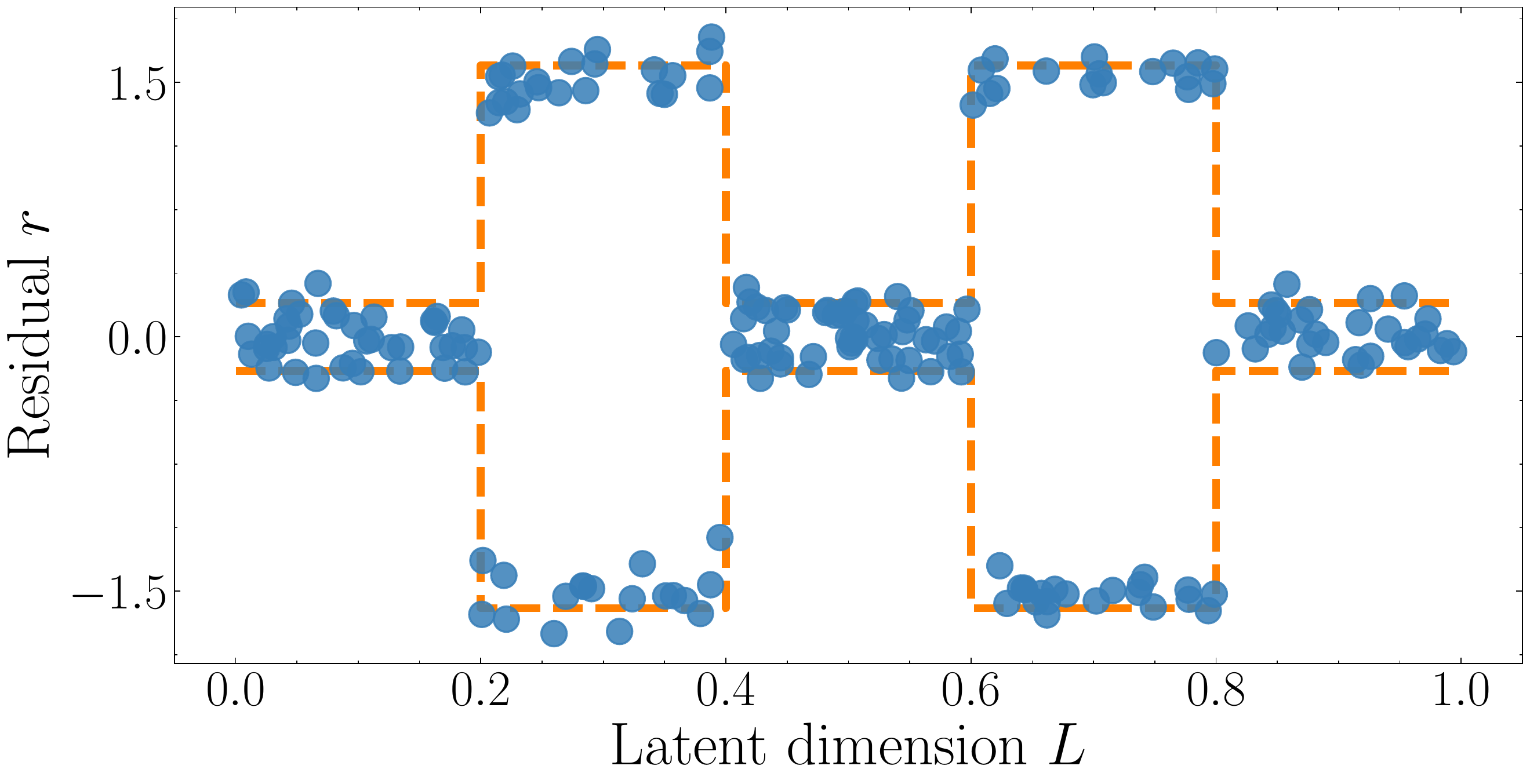}
\caption{200 observations drawn from our synthetic distribution with the distributions ground truth 90\% prediction intervals highlighted with a dashed line. }\label{fig:synth_data}
\end{figure}

\textbf{Data Generation.} In order to visualize the results, we assume both the inputs $x$ and the model residuals $r$ are generated as a function of a single, uniformly distributed latent dimension $L \in \mathcal{U}(0,1)$. We assume that there are discrete regions in which the model has made larger errors and therefore its residuals are distributed as $r \in \mathcal{N}(\text{step}(L), 0.1)$, where $\text{step}()$ denotes a step function defined as
\begin{equation}
    \text{step}(x) = \begin{cases}
  1.5 & \text{if $x \in (0.2,0.4)$ or $x \in (0.6,0.8)$}\\
  0.1 & \text{otherwise}.
\end{cases}
\end{equation}
The sign of the residual is chosen at random. A visualization of these residuals (plotted prior to taking their absolute value, which maps them to $\R^+$ as required) is provided in Fig.~\ref{fig:synth_data}.

\vspace{5mm}
\vspace{5mm}

The input features $x$ are also sampled as a function of $L$. We consider a 20-dimensional input consisting of 10 uniformly sampled pairs of points along the circumference of a circle with its base at the origin and a radius of $L$. This results in an increasingly sparse input space for larger values of $L$.

\textbf{Results.} Predicting $r$ from $x$ is challenging in the small data setting. However, by applying self-supervision to the inputs $x$, we can better learn the latent process, which in turn can provide useful information for the supervised conformal normalization model. One of the most simple self-supervised tasks is a standard autoencoder that reconstructs the input. 

Taking advantage of our knowledge of the data generation process, we consider a single latent unit in the bottleneck layer. Due to the increasing sparsity of inputs with increasing $L$, this would result in self-supervised errors that are linearly correlated with $L$ (i.e. the autoencoder could reasonably model the single ground truth latent dimension $L$). Since the residuals are a function of $L$, this self-supervised error provides a useful feature. 

\begin{figure}[h]
\includegraphics[width=0.99\linewidth]{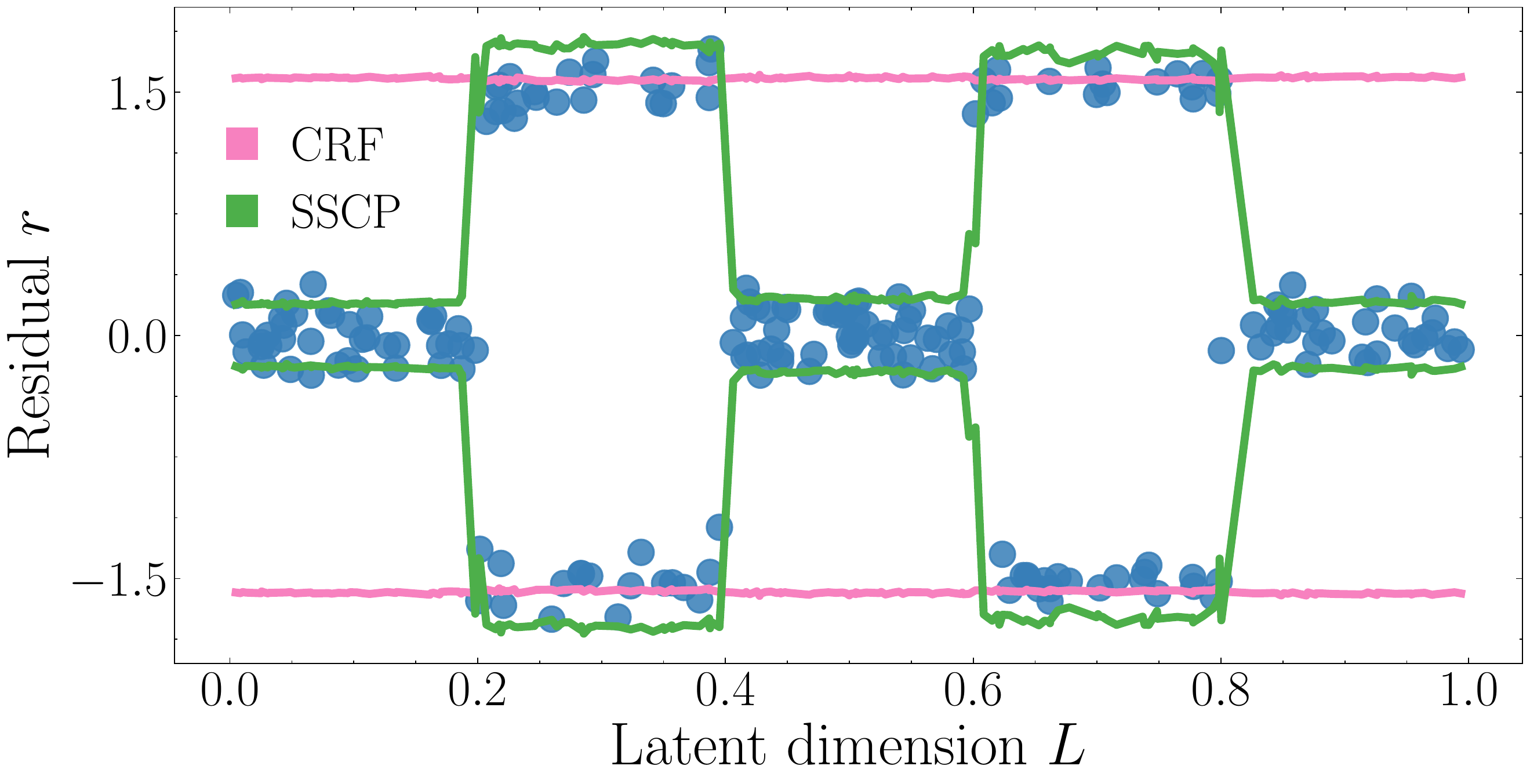}
\caption{\textbf{Conformal prediction intervals.} Standard CRF (pink) fails on this synthetic task due to a complex relationship between the inputs and the latent dimension. By including the errors of a self-supervised task as an input, SSCP (green) is able to learn much more efficient prediction intervals.
}\label{fig:synth_results}
\end{figure}

In Fig.~\ref{fig:synth_results} we compare the conformal prediction intervals ($\alpha=0.1$) of a random forest of 1000 trees with and without the additional self-supervised feature, where out-of-bag predictions are used for calibration. 

We observe that while both methods achieve the desired level of marginal coverage, the additional information from the self-supervised input results in adaptive intervals. However, the standard approach is unable to model the relationship between inputs and residuals, and is forced to default to more conservative intervals in order to maintain coverage.

This example illustrates that, for a given sample, strictly narrower intervals might not be optimal. Indeed, for samples with large residuals, CRF achieves narrower intervals than SSCP. However, these intervals are overly tight, and coverage is only maintained over the entire dataset due to overly conservative coverage in low residual regions.

\vspace{2mm}
\vspace{2mm}

\textbf{\textcolor{BrickRed}{Takeaway.}}
Augmenting the conformal normalization model's inputs with a self-supervised loss better models the heteroscedasticity in the predictive model's residuals, which results in more adaptive conformal prediction intervals. 

\subsection{Real data} \label{real}

We now assess the performance of SSCP on multiple real-world regression datasets. These datasets have been used in \cite{romano2019conformalized}. We assess the utility of using self-supervision in two plausible scenarios.

\subsubsection{Fully labeled data setting}\label{label-exp}

\textbf{Setup.} This experiment evaluates the utility of SSCP if we \textit{only} have access to fully labeled data (i.e. only the data used to train the predictive model). We evaluate whether the same labeled data can be repurposed for self-supervision simply by ignoring the label (i.e. we have no unlabeled data in $\mathcal{D}_{train}$). The benefit of this setup is that it is applicable to any supervised learning setting. Formally, $\mathcal{D}_{labeled}$ is split as follows $\mathcal{D}_{labeled}= \{\mathcal{D}_{train} \cup \mathcal{D}_{cal} \cup \mathcal{D}_{res} \cup \mathcal{D}_{test} \}$. 

The datasets as described in Sec. \ref{method} are used for both the predictive and normalization models. We can use $\mathcal{D}_{train}$ for the self-supervised tasks, simply by ignoring the label.

\textbf{Analysis.} We assess the aforementioned setup on multiple real-world datasets. Table \ref{cfr-semi-tabular} highlights that indeed the SSL task on labeled data improves the conformal prediction intervals across all datasets in terms of reduced width and excess, occasionally at the expense of slightly larger deficit. However, coverage is maintained in all scenarios. 

We also note that, of course, the smaller datasets (with respect to samples) see the smallest improvements when compared to those datasets where we have more samples. This is expected, as we simply have fewer data samples on which the SSL task can learn. That said, irrespective of the number of samples, there is still a benefit in repurposing existing labeled data to extract self-supervised information.

\vspace{+3mm}
\begin{table}[!h]
\centering
\caption{Assessing the impact of the SSL task on improving CRF, even \emph{without} additional data. We find in all cases, averaged over 5 runs, that the additional SSL information can help to improve width, deficit, and excess ($\downarrow$ better).}
\scalebox{0.95}{
\begin{tabular}{ccccc}
\toprule
    Dataset & Metric & CRF & SSCP \\ \midrule
    \multirow{3}{*}{concrete (n=1030)} 
    & Avg.Width & 0.768  & \bf 0.743 \\
    & Avg.Deficit & 0.099 & \bf 0.098 \\
    & Avg.Excess & 0.263  & \bf 0.253 \\
    \midrule

    \multirow{3}{*}{community (n=1994)}  
    & Avg.Width & 2.602  & \bf 2.462 \\
    & Avg.Deficit & 0.355  & \bf 0.361 \\
    & Avg.Excess & 1.030  & \bf 0.967 \\ 
    \midrule

    \multirow{3}{*}{star (n=2161)}  
    & Avg.Width & 0.293  & \bf 0.263 \\
    & Avg.Deficit & 0.036 & \bf 0.031 \\
    & Avg.Excess & 0.120  & \bf 0.100 \\
    \midrule

    \multirow{3}{*}{bike (n=10886)} 
    & Avg.Width & 0.720  & \bf 0.690 \\
    & Avg.Deficit & 0.164  & \bf 0.161 \\
    & Avg.Excess & 0.244  & \bf 0.232 \\
    \midrule     
  
    \multirow{3}{*}{Blog data (n=52397)} 
    & Avg.Width & 3.474  & \bf 3.360 \\
    & Avg.Deficit & \bf 3.084  & 3.155 \\
    & Avg.Excess & 1.292  & \bf 1.227\\
    \midrule

    \multirow{3}{*}{Facebook (n=81311)} 
    & Avg.Width & 1.917  & \bf 1.860 \\
    & Avg.Deficit & \bf 1.956  & 1.998 \\
    & Avg.Excess & 0.584  & \bf 0.554 \\
     
\bottomrule
\end{tabular}}
\label{cfr-semi-tabular}
\end{table}
\vspace{3mm}

\textbf{\textcolor{BrickRed}{Takeaway.}} Self-supervision improves the quality of conformal prediction intervals, even on a fixed dataset without additional data, simply by repurposing the labeled data.

\subsubsection{Augmented with unlabeled data setting}\label{unlabel-exp}

\textbf{Setup.} This experiment highlights the value of self-supervision, coupled with unlabeled data, to improve CRF-based models. We consider the typical setup in self-supervised learning, with $\mathcal{D}_{labeled}$ and $\mathcal{D}_{unlabeled}$. 

We evaluate the effect for varying proportions $p$ of labeled examples. 
$\mathcal{D}_{labeled}$ is then split and used for the predictive and normalization models, as in Sec. \ref{label-exp}. $\mathcal{D}_{unlabeled}$ augments the dataset used to train the self-supervised model (i.e. VIME). 

\begin{figure}
    \centering
    \includegraphics[width=0.4\textwidth]{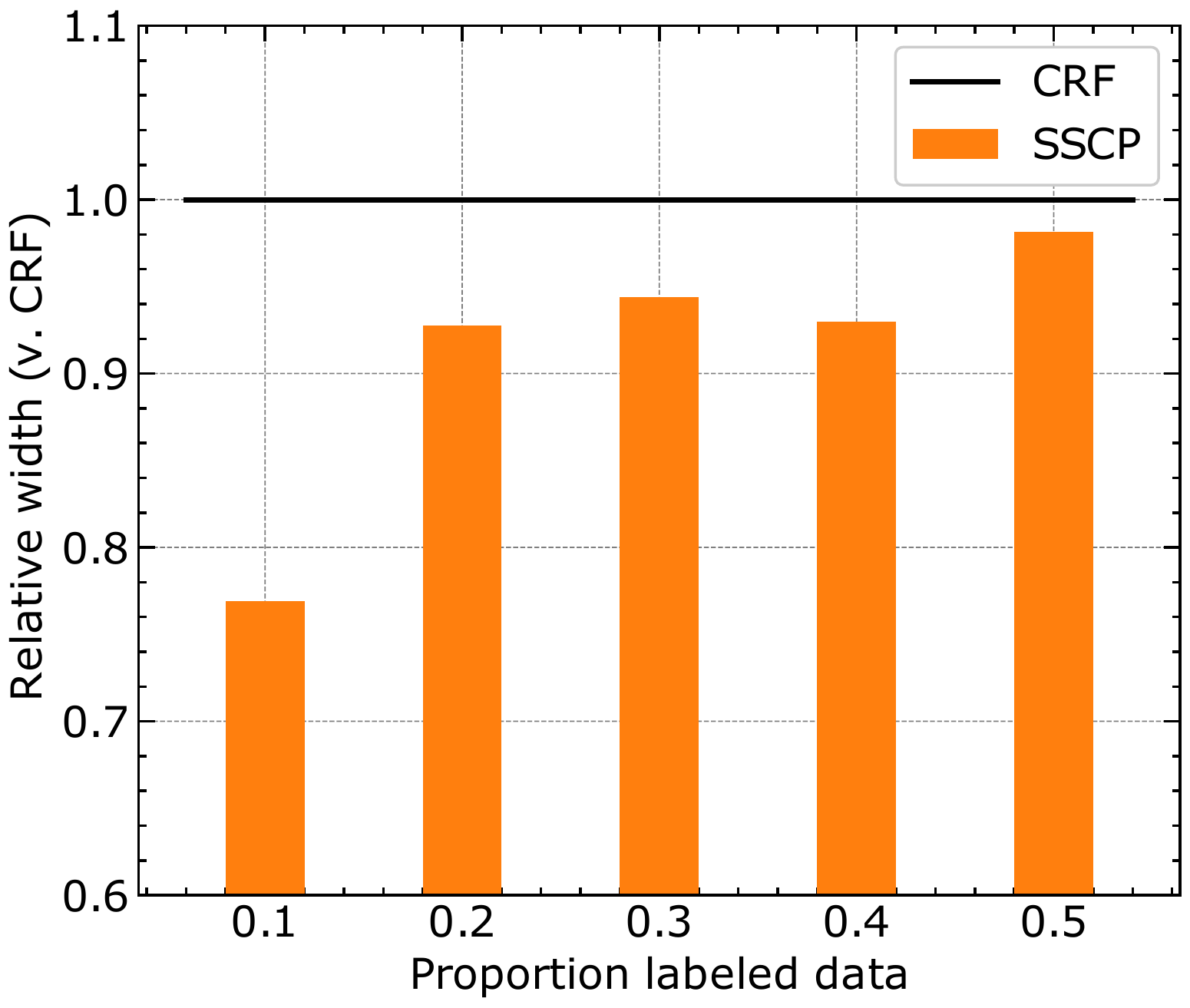}
    \caption{Relative prediction interval widths of SSCP compared to CRF for varying proportions of labeled data ($\downarrow$ better). Avg. interval width across datasets is lower with the inclusion of unlabeled data across all $p$. }
    \label{fig:unlabeled_exp}
\end{figure}

\textbf{Analysis.} What happens for varying $p$? For different values of $p$, the inclusion of the unlabeled data through self-supervision reduces the width of the prediction intervals (Fig.~\ref{fig:unlabeled_exp}).  Of course, as $p$ increases, so too does the quality of the predictive and normalization models. 

This explains the natural increase in relative width of SSCP vs. CRF for increasing $p$, i.e. as expected, with more supervised data to train on (and less unlabeled data), there is a reduced benefit of the unlabeled data. However, even as $p$ increases, we still note the benefit of the self-supervised signal in reducing the width of the intervals relative to CRF.

As an ablation, to quantify the benefit of the unlabeled data, we train the self-supervised task on ONLY the smaller $\mathcal{D}_{labeled}$, rather than the larger combination which includes $\mathcal{D}_{unlabeled}$.  The results indicate that the inclusion of unlabeled data is indeed beneficial with tighter intervals. These results can be found in Appendix \ref{ablation-unlabel}.

Finally, we expect that the choice and nature of the SSL task is impactful, and specific pretext tasks might provide superior signal to the normalization model. We discuss this in greater detail in Appendix \ref{other-ssl} by assessing alternative modes of self-supervision.

\textbf{\textcolor{BrickRed}{Takeaway.}} Self-supervision using unlabeled data helps to improve the quality of conformal intervals, across different proportions of labeled vs. unlabeled data.

\subsection{Insights}\label{insights-exp}
\begin{figure}[!h]
\centering
    \begin{subfigure}[t]{.23\textwidth}
        \includegraphics[width=\textwidth]{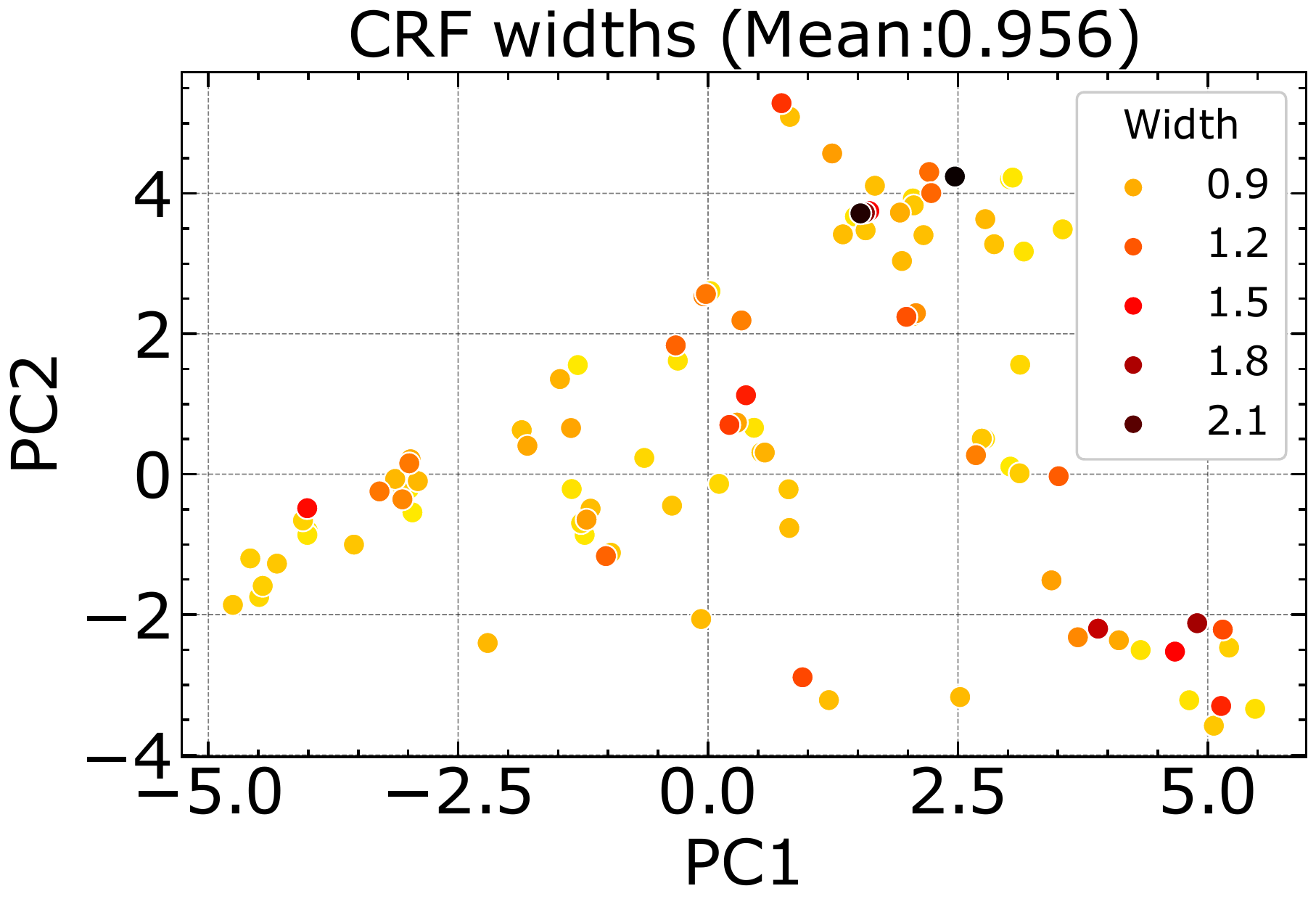}
        \caption{\footnotesize{CRF}}
    \end{subfigure}%
    ~
    \begin{subfigure}[t]{.23\textwidth}
        \includegraphics[width=\textwidth]{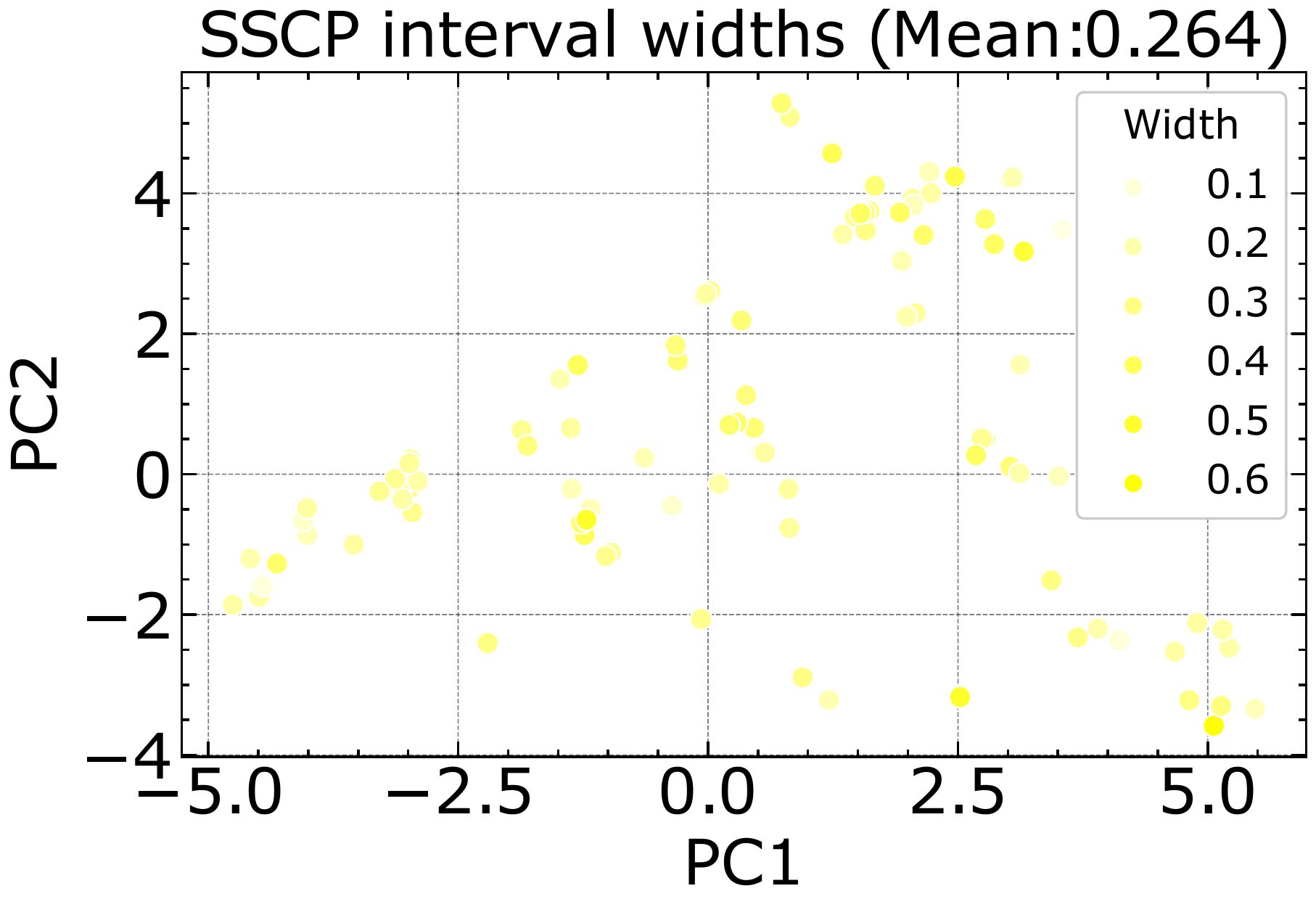}
        \caption{\footnotesize{SSCP}}
    \end{subfigure}%
    \label{pca}
    \vspace{-3mm}
    \caption{\footnotesize{Prediction interval width for most challenging samples on Star. SSCP reduces interval width }}\label{fig:insight_pca}
\end{figure}

Let us now examine where SSCP helps the most. First, we compare the prediction intervals of samples with CRF vs. the same samples with SSCP. We find that the main source of improvement over CRF is for those samples with the largest interval width. Fig.~\ref{fig:insight_pca} highlights that SSCP significantly reduces the interval width for these samples, which is the key source of the reduction in average width.

This begs the question on what type of samples/regions SSCP helps? To facilitate visualization, we reduce the dimensionality of the input features $\mathbf{X}$ using principal component analysis (PCA). We plot PC1 vs. output $y$ and compare the prediction intervals for SSCP vs. CRF. 

\begin{figure}[!h]
\centering
    \begin{subfigure}[t]{.23\textwidth}
        \includegraphics[width=\textwidth]{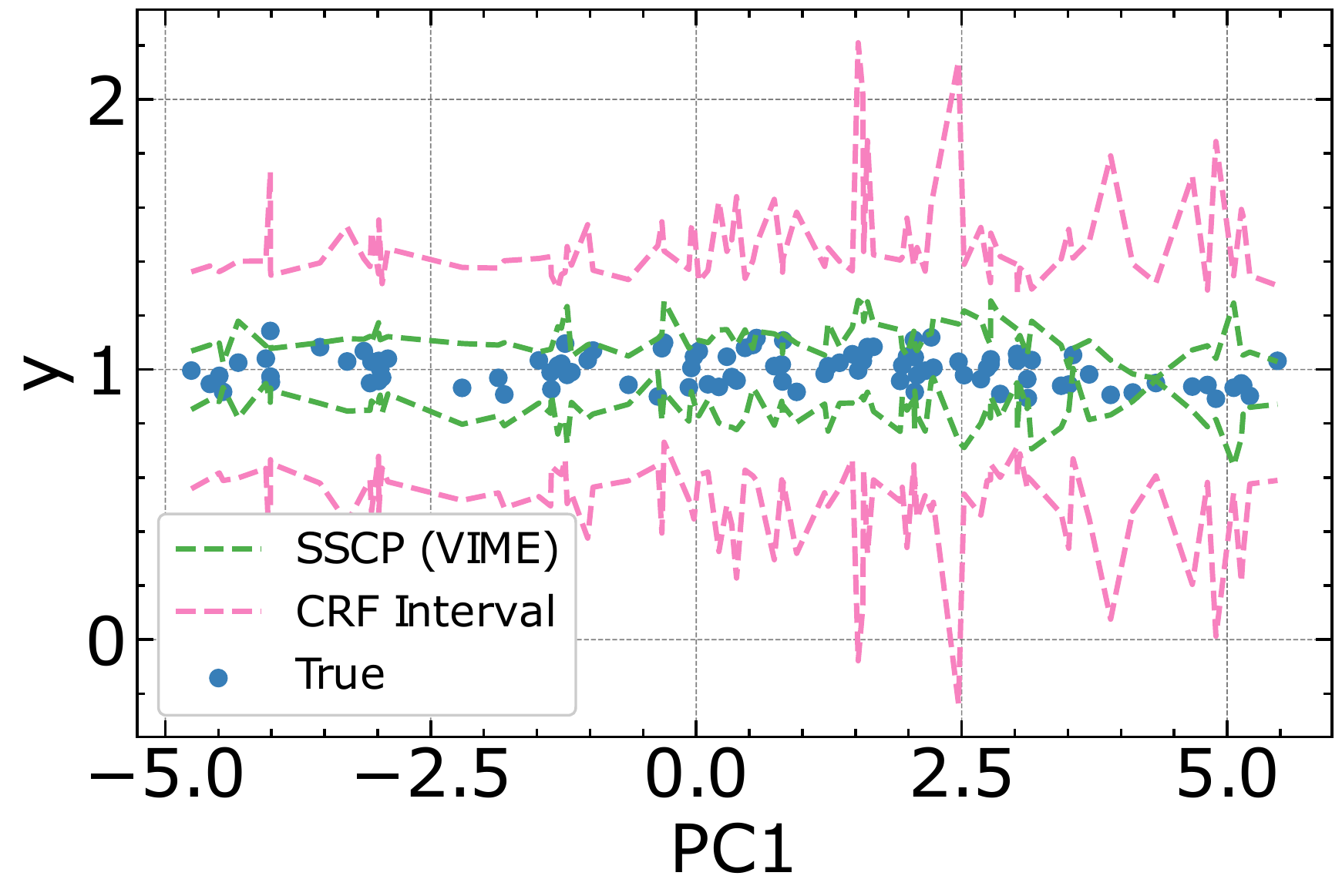}
        \caption{\footnotesize{Star dataset}}
    \end{subfigure}%
    ~
    \begin{subfigure}[t]{.25\textwidth}
        \includegraphics[width=\textwidth]{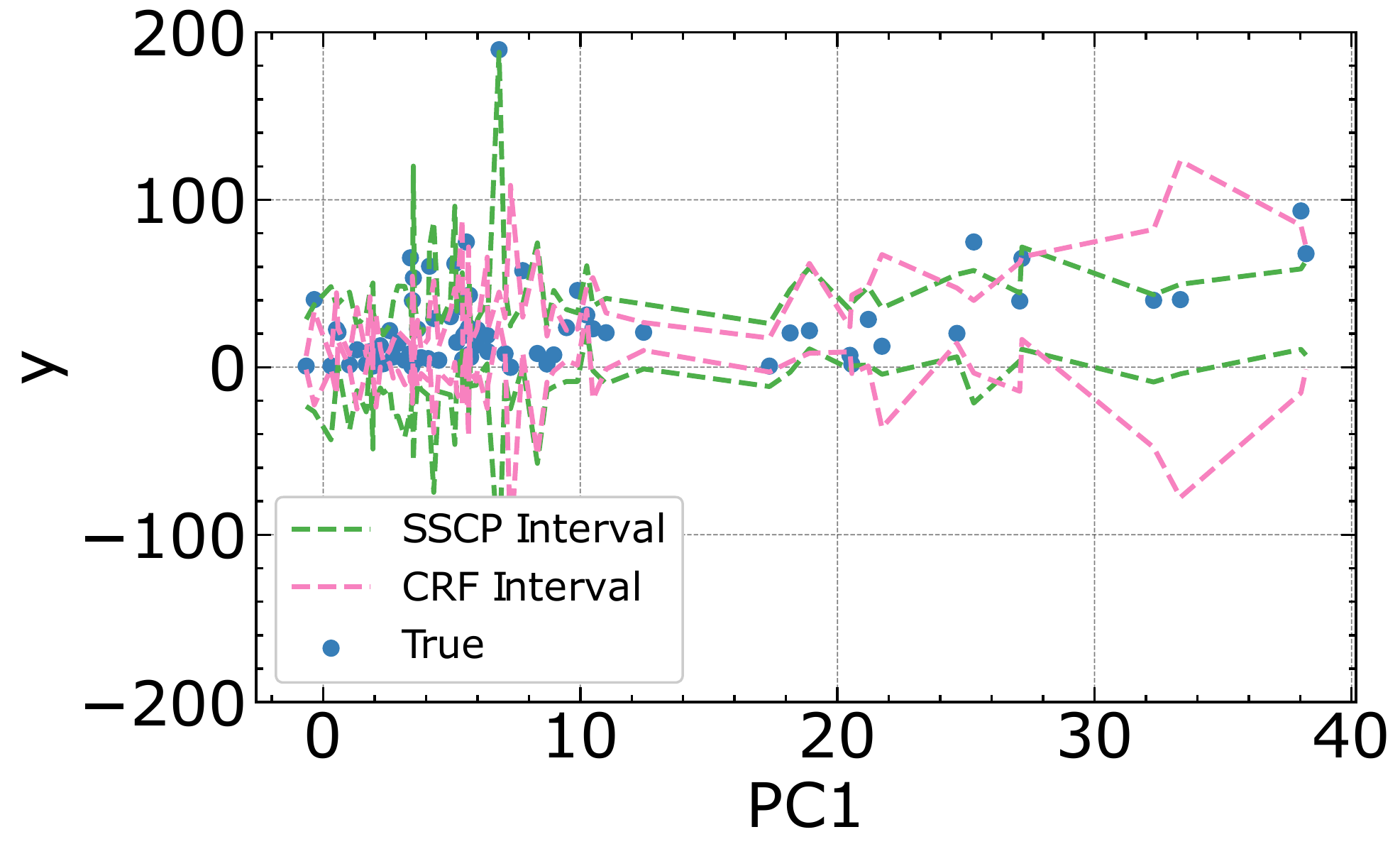}
        \caption{\footnotesize{Facebook dataset}}
    \end{subfigure}%
     \vspace{-3mm}
    \caption{\footnotesize{Comparing the prediction intervals in different regions. SSCP provides benefit in the sparser regions.}}\label{fig:insight_plot}
\end{figure}

In Fig.~\ref{fig:insight_plot}, we examine two coarse regions of the PC1 space: dense and sparse. In the denser regions, CRF has narrower intervals, similar to SSCP. Whilst, in the sparser regions, CRF often has much wider prediction intervals compared to SSCP. It is in these sparse regions that SSCP clearly helps.

The rationale is that in the sparser regions, the conformal normalization model $\sigma$ does not perform well. Hence, to satisfy the coverage constraint, the CRF intervals are more conservative and therefore more inefficient in these challenging sparser regions.

In contrast, SSCP with the SS signal helps to improve the conformal normalization model (i.e. the residuals are predicted more accurately) and hence by virtue of the improved $\sigma$ it achieves intervals with lower width on the sparser challenging regions. Quantitatively, this improvement to $\sigma$ in MAE is $\pm 14\%$ and $\pm 5\%$ for Fig.~\ref{fig:insight_plot} (a) and (b) respectively. Further analysis is found in Appendix \ref{deep-dive} and \ref{robustness-study}.

\textbf{\textcolor{BrickRed}{Takeaway.}} SSCP improves prediction interval width on the most challenging samples and sparser regions.

\section{DISCUSSION}
Improving the quality of prediction intervals is crucial for more reliable uncertainty quantification. In this work, we show the value of self-supervised learning beyond model pretraining and representation learning. We propose a framework to integrate self-supervised losses to improve conformal prediction intervals. We empirically show on both synthetic and real data that self-supervision improves the prediction intervals of locally adaptive conformal prediction, specifically CRF. In addition, the self-supervision assists in more challenging and sparser regions.

\textbf{Limitations.} SSCP provides a framework to highlight the value of self-supervision for conformal prediction. Our evaluation of pretext tasks (in the main paper and appendix) are provided as examples and additional tasks could be evaluated using the provided ``recipe'', inclusive of future work to design a conformal prediction-specific pretext task. 

\textbf{Societal impact.} Quantifying predictive uncertainty is important for ML, especially in high-stakes settings such as medicine and finance. SSCP seeks to improve this and provide more trustworthy predictions, by providing end users with reliable prediction intervals as a measure of confidence. 

\textbf{Future work.} The findings of this paper open up avenues for future research, leveraging the idea that self-supervision can be used to improve conformal prediction. The following are proposed areas of future exploration: (i) assessing alternative self-supervised approaches (e.g. \citealp{chen2020simple,lee2022self}), (ii) development of new conformal-aware self-supervised tasks, (iii) application of self-supervision to improve feature-wise conformal prediction (as per \citealp{seedat2022data}, \citealp{teng2022predictive}), (iv) the use of self-supervision with other data modalities such as images, where the self-supervised tasks tend to be distinct from the tabular setting. 

\section*{ACKNOWLEDGEMENTS} The authors are grateful to Alicia Curth, Boris van Breugel, Nicolas Huynh, Paulius Rauba, Tennison Liu, and the anonymous AISTATS reviewers for their comments and feedback on an earlier manuscript. Nabeel Seedat and Alan Jeffares are funded by
the Cystic Fibrosis Trust. Fergus Imrie and Mihaela van der Schaar are supported by the National Science Foundation (NSF, grant number 1722516).
Mihaela van der Schaar is additionally supported by the Office of Naval Research (ONR).

\bibliography{aistats}
\bibliographystyle{unsrtnat}

\clearpage
\onecolumn
\appendix

\section{COMMENT ON SSCP COVERAGE} \label{app:coveragecomment}

Conformal prediction was originally introduced in \cite{vovk2005algorithmic}. However, this so-called \textit{transductive} method is impractical for naive use with most modern algorithms due to multiple refits of the model being required per test point. \textit{Inductive} (or \textit{split}) conformal prediction \citep{papadopoulos2008inductive} provides a satisfying solution to this limitation by separating the fitting of the model and the calibration of the conformal predictor across two disjoint datasets. As noted in \cite{tibshirani2019conformal,zhang2020autocp}, by considering the trained model $\mu$ fixed, the non-conformity scores on the calibration and test data are clearly exchangeable provided the data itself is exchangeable. Of course, this method results in constant interval width across all test points, which is determined by the critical non-conformity score $\epsilon$.
In order to provide adaptive width intervals that reflect the difficulty of a particular example, \cite{papadopoulos2010neural} introduced a residual estimator $\sigma$ resulting in CRF. By adjusting the denominator of the non-conformity score function to $\gamma(x) = \frac{|y - f(x)|}{\sigma(x)}$, the prediction intervals become $(\mu(x) - \epsilon \cdot \sigma(x), \mu(x) + \epsilon \cdot \sigma(x))$ and still enjoy marginal coverage in the inductive setting. We emphasize that it is only required that the calibration data and test data are exchangeable, both $\mu$ and $\sigma$ can be trained on any disjoint data provided they are subsequently held fixed.

It is straightforward to check that the method proposed in this work (SSCP) can be viewed as a special case of CRF as described in the previous section and, therefore, maintains the same coverage guarantees. The additional self-supervised function $f_{\text{ss}}$ can simply be viewed as a sub-component of an alternative residual model, defined as $\hat{\sigma}(x) = \sigma(\{ x, f_{\text{ss}}(x))$. Just as before, and provided this model remains fixed during calibration and testing, the same conformal prediction coverage guarantees hold. 

\clearpage

\section{EXPERIMENTAL DETAILS}\label{appx-b}

In this section, we outline further details of the experimental setup used in Section \ref{experiments}.

We assess SSCP on multiple open-source regression datasets from the UCI repository \citep{uci}. These datasets vary both in the number of features as well as in the number of samples (documented in Table \ref{cfr-semi-tabular}). Similar to \cite{romano2019conformalized}, we standardize the features to have zero mean and unit variance. The output $y$ is rescaled by the mean absolute value.

We compute the performance metrics averaged over 5 random runs. In each run, we change the train-test split sampling, where the split proportion of 80-20. $\mathcal{D}_\text{res}$ is then split 80-20 with $\mathcal{D}_\text{train}$ and then subsequently $\mathcal{D}_\text{cal}$ split 80-20 with the remaining $\mathcal{D}_\text{train}$. We fix $\alpha$=0.1.

\subsection{Predictive and conformal normalization models.}
Our baseline models, both the predictive model and conformal normalization model, are implemented using neural networks. Our architectures are similar to those used in \cite{romano2019conformalized}.

The neural network is an MLP with three fully connected layers, each with ReLU activations. The input is an $n$-dimensional feature vector $X$ passed to an input layer with 64 hidden units. The hidden layer also contains 64 units. Finally, the output layer is a single neuron corresponding to the regression output $y$. 

Mean squared error loss is used throughout our experiments, optimized using Adam with a learning rate of 5e-4. Batch size is 128 and we apply dropout regularization with dropout probability of 0.1. Finally, we train using early stopping with patience of 20.

We extract up to the second fully connected layer as the encoder $e$.

\subsection{Self-supervised learning}

\paragraph{VIME.} We specifically leverage the self-supervised component of VIME. This consists of an encoder together with an output mask estimator and feature estimator, which are jointly optimized. We keep the default architecture as per \cite{yoon2020vime} with a dense layer for each. 

We train with a batch size of 128 for 500 epochs with early stopping enabled. The probability of corruption used is 0.3 and the weight of the feature to mask loss $\alpha$ is set to 2.
Future work could assess how to optimally set these parameters specific to the conformal prediction setting.

A minor difference to standard VIME in our application is that rather than the input being raw features, we provide the output from encoder $e$ as the task input, and train VIME as the self-supervised predictor $h$.

\paragraph{Autoencoder.} Similar to VIME rather than the input being raw features, we provide the output from the encoder $e$ as the input, and simply train the autoencoder decoder portion as the self-supervised predictor $h$.

The decoder has the same layers and dimensionality as the encoder $e$, simply being reversed such that the reconstructed output matches the dimensionality of the raw features.

We train with a batch size of 128 for 500 epochs with early stopping enabled. MSE is the loss function. Similar to VIME, the nature and depth of the autoencoder could be assessed as future work concerning how to optimally set these parameters specific to the conformal prediction setting.

\clearpage

\section{ADDITIONAL EXPERIMENTS}\label{appx-c}

\subsection{Ablation: unlabeled data experiment}\label{ablation-unlabel}

\paragraph{Goal.}
This experiment seeks to understand the benefit of the unlabeled data to SSCP. To do so, we perform an ablation of the unlabeled data when used with SSCP for varying proportions $p$ of labeled data. In the ablation variant, i.e. SSCP (Labeled), we only re-use the fully labeled data for the self-supervised task and compare it to SSCP (ALL) which augments the self-supervised dataset with the unlabeled data. 

\paragraph{Analysis.} The results outlined in Table \ref{tab:ablate2} highlight the benefit of the unlabeled data, especially true in the low data regime when the labeled dataset is small (i.e. low p),  where the ablation SSCP (Labeled) showcases an increased average interval width, i.e. when not including the unlabeled data.

\begin{table}[!h]
    \centering
    \caption{Avg. width of intervals with and without unlabeled data ($\downarrow$ is better)}
    \begin{tabular}{c|c|c|c}
    p & SSCP (ALL) & SSCP (Labeled) & Difference \\\hline
0.1 & 3.603 & 3.893 & 0.290 \\ \midrule
0.2 & 2.878 & 3.000 & 0.122 \\ \midrule
0.3 & 1.796 & 1.752 & -0.044 \\ \midrule
0.4 & 1.505 & 1.459 & -0.045 \\ \midrule
0.5 & 1.465 & 1.490 & 0.025 \\ \midrule
    \end{tabular}

    \label{tab:ablate2}
\end{table}

\subsection{Sanity check: alternative sources of signal}

\paragraph{Goal.}
We aim to assess the value of alternative sources of signal with respect to the residual prediction, in order to highlight the value imbued by the self-supervision. We assess (i) using an isolation forest instead of SSL and (ii) using the self-supervised signal directly, rather than as input to a residual/normalization model $\sigma$.

\paragraph{Analysis.} 
Sanity check comparing alternative self-supervised signals to SSCP (VIME). We analyze adding an isolation forest in place of VIME i.e. SSCP (IF) and simply using the SSL signal for normalization, instead of training $\sigma$ (SSL-normalizer). The results are shown in Table \ref{sanity-check} and indicate that indeed overall the usage of VIME provides the best intervals in the most stable manner.

\begin{table}[!h]
\centering
\caption{Sanity check comparing alternative self-supervised signals to SSCP (VIME). We report mean average interval width over 5 runs. ($\downarrow$ better) }
\scalebox{1}{
\begin{tabular}{cccc}
\toprule
    Dataset  & SSCP (VIME) & SSCP (IF) & SSCP (SSL-normalizer) \\ \midrule

     \multirow{1}{*}{Overall} 
     & \bf 1.587 & 1.652 & 4.139 \\
     \midrule

    \multirow{1}{*}{concrete(n=1030)} 
    & \bf 0.743 & 0.904 & 0.761 \\
     \midrule

    \multirow{1}{*}{community(n=1994)} 
   & \bf 2.462 & 2.581 & 2.468 \\ 
     \midrule
     
    \multirow{1}{*}{star(n=2161)} 
    &   0.263 & 0.230 & \bf 0.207 \\
     \midrule
     
    \multirow{1}{*}{bike(n=10886)} 
   & \bf  \bf 0.690 & 0.711 & 1.016 \\
  \midrule     

     \multirow{1}{*}{Blog data (n=52397)} 
 & \bf 3.360 & \bf 3.360 & 11.823 \\   \midrule

     \multirow{1}{*}{Facebook (n=81311)} 
  & \bf 1.860 & 2.127 & 8.562
 \\
     \midrule

     \midrule
\bottomrule
\end{tabular}}
\label{sanity-check}
\end{table}

\newpage

\subsection{Conformalized Quantile Regression} \label{app:CQR}
Conformalized Quantile Regression (CQR) \citep{romano2019conformalized} provides a powerful alternative to CRF for achieving locally adaptive intervals for conformal prediction. The method builds upon conditional quantile regression \citep{koenker1978regression} in which the pinball loss objective is utilized such that a model predicts upper and lower quantiles around a target by enforcing non-symmetric and opposite loss functions for the two model outputs. This method was then conformalized in \cite{romano2019conformalized} using an inductive conformal prediction procedure. 

Although CQR removes the flexibility of post hoc conformalization of an arbitrary model, it often benefits from more efficient intervals in practice. Therefore, one might wish to consider it as an alternative to CRF in our SSCP framework. The key difference between the two methods is that CQR no longer requires a separate conformal normalizer model, as the prediction intervals $[l(x), r(x)]$ are derived from the main model's quantile predictions. Instead, we pass the self-supervised errors as an additional input into the encoder of the main model. Sharing the encoder between the self-supervised model $f_{ss}$ and the main model $f$ (i.e. $e = e_{ss}$) can still be achieved by passing a placeholder (0) for the additional feature during the self-supervised phase. In our experiments, we consider both a shared encoder between the two models (Shared) and an independent encoder per model (Indep). This procedure is summarized in Algorithm \ref{alg:CQRSSCP}. 
\begin{algorithm}
\caption{Self-supervised Conformal Prediction with CQR}
\begin{algorithmic}[1] 
\State \textbf{Input:} $\mathcal{D}_\text{train}$, $\mathcal{D}_\text{cal}$, $e$, $e_{ss}$, $g$, $h$
\Procedure{SSCP}{}      
    \State Train $f_{ss} = h \circ e_{ss}$ on  $\mathcal{D}_\text{train}$ \;
    \State Calculate self-supervised errors $\mathcal{L}_{\text{ss}}^\text{train}$ on $\mathcal{D}_\text{train}$ and $\mathcal{L}_{\text{ss}}^\text{cal}$ on $\mathcal{D}_\text{cal}$ \;
    \State Train $f = g \circ e$ on  $\mathcal{D}_\text{train} \cup \mathcal{L}_{\text{ss}}^\text{train}$  \;
    \State Apply standard CQR calibration on $\mathcal{D}_\text{cal} \cup \mathcal{L}_{\text{ss}}^\text{cal}$ \;
    \State \textbf{Return:} $e$, $e_{ss}$, $f$, $g$, $h$ for test evaluation \;

\EndProcedure
\end{algorithmic}
\label{alg:CQRSSCP}
\end{algorithm}

\begin{figure}
    \centering
\includegraphics[width=0.4\textwidth]{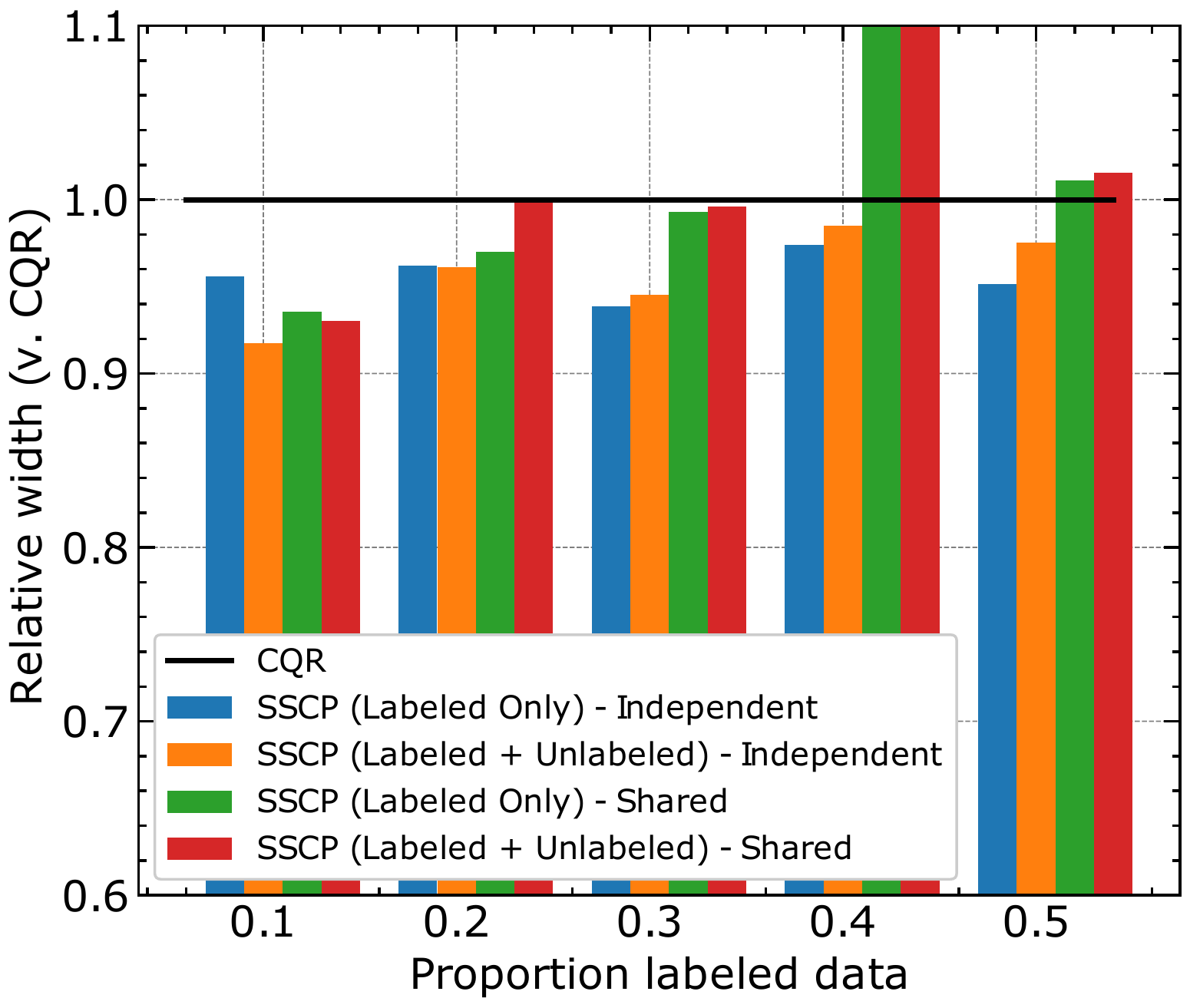}
    \caption{Repeating the partially labeled experiments from Section \ref{unlabel-exp} with CQR. Evaluating independent vs. shared representations and the inclusion of additional unlabeled data for various proportions of labeled data. Lower proportions of labeled data achieve the largest gains over standard CQR.}
    \label{fig:cqr_unlabel}
\end{figure}

We include the results of applying SSCP with CQR to the same experiments described in Sections \ref{label-exp} \& \ref{unlabel-exp} in Table \ref{cqr-labeled} \& Figure \ref{fig:cqr_unlabel} respectively. In the fully labeled setting we find that (1) independent encoders outperform shared encoders using the method we describe, and (2) CQR augmented with a self-supervised signal is competitive with and often improves vanilla CQR in terms of average interval width. In the partially labeled setting, we observe a similar trend that for lower quantities of labeled data, SSCP provides the largest gains over standard CQR. The trend between the four methods indicates, consistent with the labeled setting, that independent encoders outperform a shared representation. However, evidence for improved performance with additional unlabeled data is inconclusive in these experiments. Hence, further research on more effective ways to (1) share a representation in this context and (2) integrate unlabeled data provide exciting directions for future work.

\begin{table}[!h]
\centering
\caption{Repeating the fully labeled experiments from Section \ref{label-exp} with CQR. We report mean average interval width ($\downarrow$ better) over 5 runs. These results indicate that using CQR augmented with a self-supervised loss (SSCP-CQR), often improves vanilla CQR.}
\scalebox{1}{
\begin{tabular}{cccc}
\toprule
    Dataset  & CQR & SSCP-CQR (Indep) & SSCP-CQR (Shared) \\ \midrule

     \multirow{1}{*}{Overall} 
     & 0.904 & \bf 0.866 & 0.894 \\
     \midrule
        
    \multirow{1}{*}{concrete(n=1030)} 
     & 0.532 & \bf  0.505 & 0.512 \\
     \midrule

    \multirow{1}{*}{community(n=1994)} 
   & 1.733 & \bf  1.603 & 1.685 \\ 
     \midrule
     
    \multirow{1}{*}{star(n=2161)} 
    & \bf  0.196 & 0.198  & 0.208 \\
     \midrule
     
    \multirow{1}{*}{bike(n=10886)} 
   & \bf  0.620 & 0.623 & 0.650 \\
  \midrule     
    
     \multirow{1}{*}{Blog data (n=52397)} 
 & 1.441 & \bf  1.402 & 1.405 \\   \midrule

     \multirow{1}{*}{Facebook (n=81311)} 
  & 1.274 & \bf 1.220 & 1.292
 \\
     \midrule

     \midrule
\bottomrule
\end{tabular}}
\label{cqr-labeled}
\end{table}

\subsection{Evaluating alternative self-supervised tasks \& the source of gain}\label{other-ssl}

\subsubsection{High-level}

\paragraph{Goal.} In the main manuscript, we have examined VIME as an example self-supervised task coupled with SSCP. We extend this analysis and explore a reconstructing autoencoder (AE) as an alternative self-supervised task. We refer to these as SSCP (AE) and SSCP (VIME).

\paragraph{Analysis.} 
We compare the performance of SSCP (VIME) and SSCP (AE) to CRF in Table \ref{cfr-compare-ae-vime-tabular}. We find that, in general, the addition of the self-supervised loss is a useful feature, resulting in more adaptive intervals. 

That being said, VIME seems to provide greater performance improvements across a variety of datasets when compared to the AE. This is of course reasonable, as VIME has been shown to be a more performant self-supervised task in general for tabular data. Next, we take a deep-dive to understand the impact of the self-supervised task and where the improvements lie.

\begin{table}[!h]
\centering
\caption{Assessing the impact of the SSL task on improving CRF, for two different self-supervised tasks (i.e. VIME and AE).  We find in general that the self-supervision helps with VIME being more performant than AE as the self-supervised task.  ($\downarrow$ better) }
\scalebox{0.85}{
\begin{tabular}{ccccc}
\toprule
    Dataset & Method & CRF & SSCP(AE) & SSCP(VIME) \\ \midrule
        
    \multirow{5}{*}{concrete(n=1030)} 
    & Avg.Width & 0.768 & 0.794 & \bf 0.743 \textcolor{ForestGreen}{$\downarrow$} \\
    & Avg.Deficit & 0.099 & 0.117 & \bf 0.098 \textcolor{ForestGreen}{$\downarrow$} \\
    & Avg.Excess & 0.263 & 0.276 & \bf 0.253 \textcolor{ForestGreen}{$\downarrow$} \\
     \midrule

    \multirow{5}{*}{community(n=1994)} 
     & Avg.Width & 2.602 &  2.667  & \bf 2.462 \textcolor{ForestGreen}{$\downarrow$} \\
    & Avg.Deficit & 0.355 &  0.382  & \bf 0.361  \textcolor{ForestGreen}{$\downarrow$}\\
    & Avg.Excess & 1.030 &  1.068  & \bf 0.967 \textcolor{ForestGreen}{$\downarrow$} \\ 
     \midrule

    \multirow{5}{*}{star(n=2161)} 
    & Avg.Width & 0.293 & \bf 0.261  \textcolor{ForestGreen}{$\downarrow$} &  0.263 \textcolor{ForestGreen}{$\downarrow$} \\
    & Avg.Deficit & 0.036 & 0.040 & \bf 0.031 \textcolor{ForestGreen}{$\downarrow$} \\
    & Avg.Excess & 0.120 & 0.110  \textcolor{ForestGreen}{$\downarrow$} & \bf 0.100 \textcolor{ForestGreen}{$\downarrow$} \\
     \midrule

    \multirow{5}{*}{bike(n=10886)} 
    & Avg.Width & 0.720 & 0.706 \textcolor{ForestGreen}{$\downarrow$} & \bf 0.690 \textcolor{ForestGreen}{$\downarrow$} \\
    & Avg.Deficit & 0.164 & 0.161  \textcolor{ForestGreen}{$\downarrow$} & \bf 0.161 \textcolor{ForestGreen}{$\downarrow$} \\
    & Avg.Excess & 0.244 & 0.241 \textcolor{ForestGreen}{$\downarrow$} & \bf 0.232 \textcolor{ForestGreen}{$\downarrow$} \\
  \midrule

     \multirow{5}{*}{Blog data (n=52397)} 
 & Avg.Width & 3.474 & 3.400 \textcolor{ForestGreen}{$\downarrow$} & \bf 3.360 \textcolor{ForestGreen}{$\downarrow$} \\
& Avg.Deficit & \bf 3.084 &  3.095  & 3.155 \\
& Avg.Excess & 1.292 & 1.246 \textcolor{ForestGreen}{$\downarrow$} & \bf 1.227 \textcolor{ForestGreen}{$\downarrow$} \\     \midrule

     \multirow{5}{*}{Facebook (n=81311)} 
  & Avg.Width & 1.917 & 1.872  \textcolor{ForestGreen}{$\downarrow$} & \bf 1.860 \textcolor{ForestGreen}{$\downarrow$} \\
& Avg.Deficit & \bf 1.956 &  2.010  & 1.998 \\
& Avg.Excess & 0.584 & 0.563  \textcolor{ForestGreen}{$\downarrow$} & \bf 0.554 \textcolor{ForestGreen}{$\downarrow$} \\
     \midrule

     \midrule
\bottomrule
\end{tabular}}
\label{cfr-compare-ae-vime-tabular}
\end{table}

\clearpage

\subsubsection{Deep-dive} \label{deep-dive}

\paragraph{Goal.} We now take a deep-dive to better understand the differences for SSCP with different self-supervised tasks, SSCP (VIME) and SSCP (AE). In addition, we seek to understand the gains from SSCP better. The analysis in this section is performed on the Star dataset.

\begin{figure}[!h]
    \centering
    \includegraphics[width=0.4\textwidth]{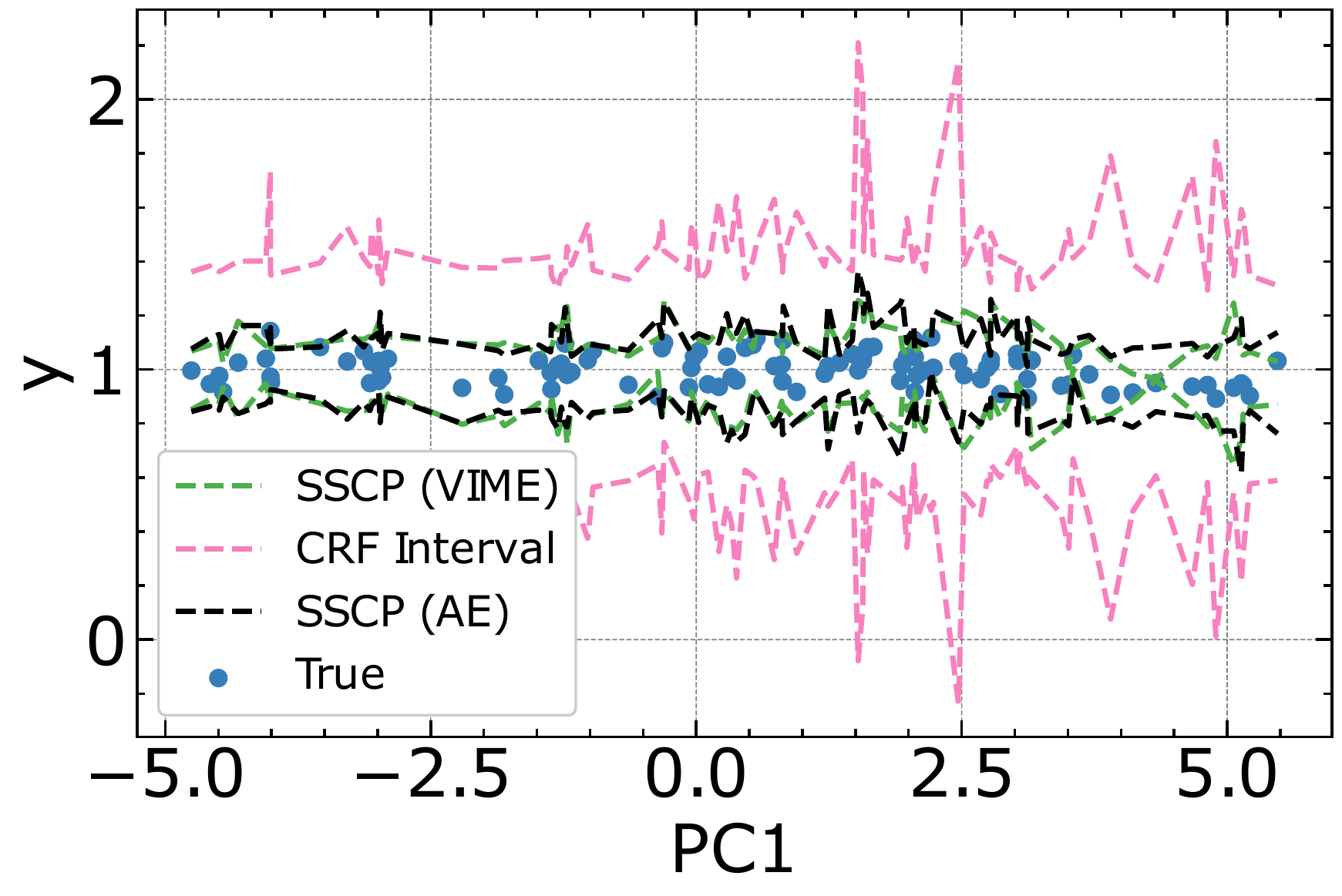}
    \caption{Comparing the prediction intervals in different regions.
All variants of SSCP provide benefit in the sparser regions. The intervals are similar for SSCP (VIME) and SSCP (AE) in the denser regions, with SSCP (VIME) generally performing better. This is especially the case in the sparser regions with narrower intervals.}
    \label{fig:intervals_all}
\end{figure}

\paragraph{Analysis.} We first compare the prediction intervals in low-dimensional space. Similar to the main manuscript, we reduce the dimensionality of the input features (X) using PCA to facilitate visualization. We plot PC1 vs. output $y$ and
compare the prediction intervals for SSCP (VIME) vs. SSCP (AE) vs. CRF. The results can be seen in Figure \ref{fig:intervals_all}.

We see that, in general, CRF results in wider intervals compared to the SSCP variants. Additionally, SSCP (VIME) generally outperforms SSCP (AE) with narrower intervals.
In the analysis that follows, we show that the majority of this gain in performance is achieved on the most uncertain examples in the more sparsely populated regions of input space. The implication is that the self-supervised loss is useful in informing the interval widths on the most uncertain examples where the highest inefficiency occurs.

We also seek to better understand why SSCP (VIME) outperforms SSCP (AE). Beyond simply being a better self-supervised task in general \citep{yoon2020vime}, we find that VIME loss also appears to have a stronger relationship with the predictive model's errors when compared to the autoencoder, with correlations of \textbf{0.33} and \textbf{0.06} respectively. 

Delving deeper beyond the mean correlation of SS loss with the main predictive model error. We find that the largest improvements in reducing interval width occur for samples that have the largest correlation of SS loss with the predictive model error ($\pm$0.35-45). We find limited improvement for samples with the lowest correlation ($\pm$0.11-0.20), highlighting the importance of the relationship between SS loss and the predictive error itself. 

As described in the main text, a stronger relationship between the self-supervised loss and the predictive model errors should typically result in a better residual model and, in turn, more efficient prediction intervals. This result confirms that hypothesis. The broader implication is that such considerations should be kept in mind when selecting the self-supervised task. Furthermore, these findings could be leveraged in future works for designing specific self-supervised tasks in the SSCP setting such that the task should seek to explicitly increase this correspondence.

\begin{figure*}[!h]

\centering
    \begin{subfigure}[t]{.33\textwidth}
        \includegraphics[width=\textwidth]{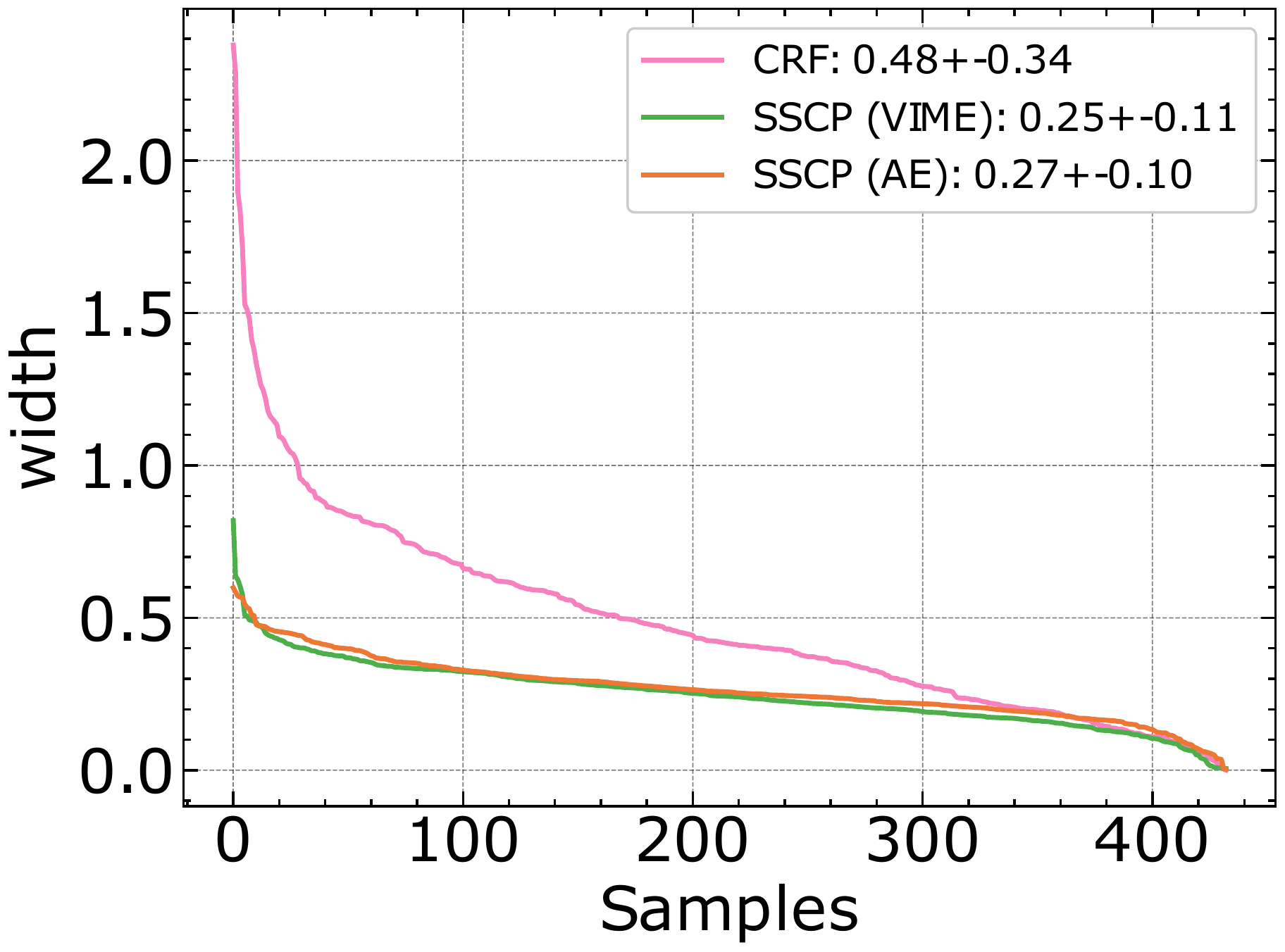}

        \caption{\footnotesize{Width}}
    \end{subfigure}%
    ~
    \begin{subfigure}[t]{.33\textwidth}
        \includegraphics[width=\textwidth]{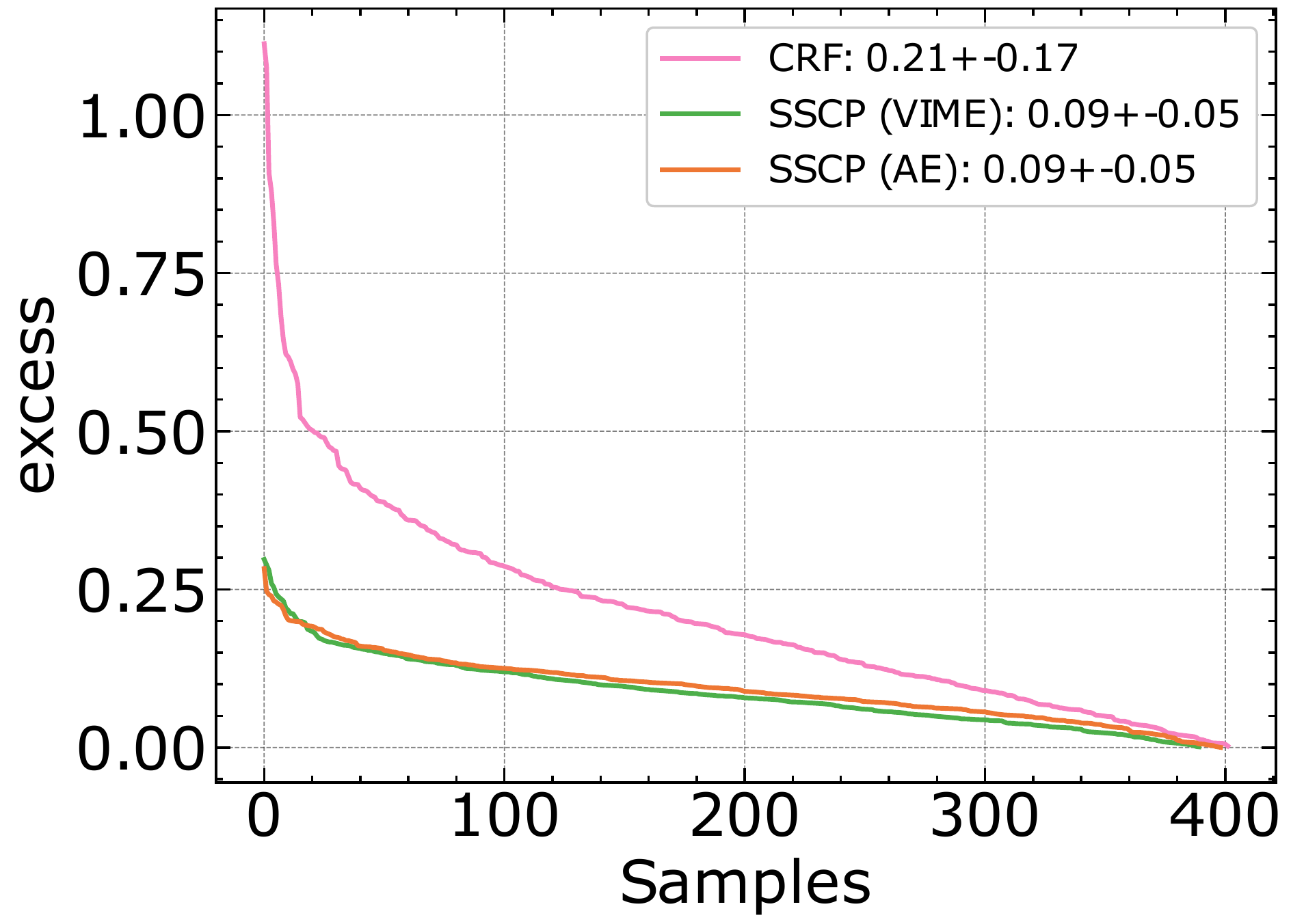}

        \caption{\footnotesize{Excess}}
    \end{subfigure}%
    ~
    \begin{subfigure}[t]{.33\textwidth}
        \includegraphics[width=\textwidth]{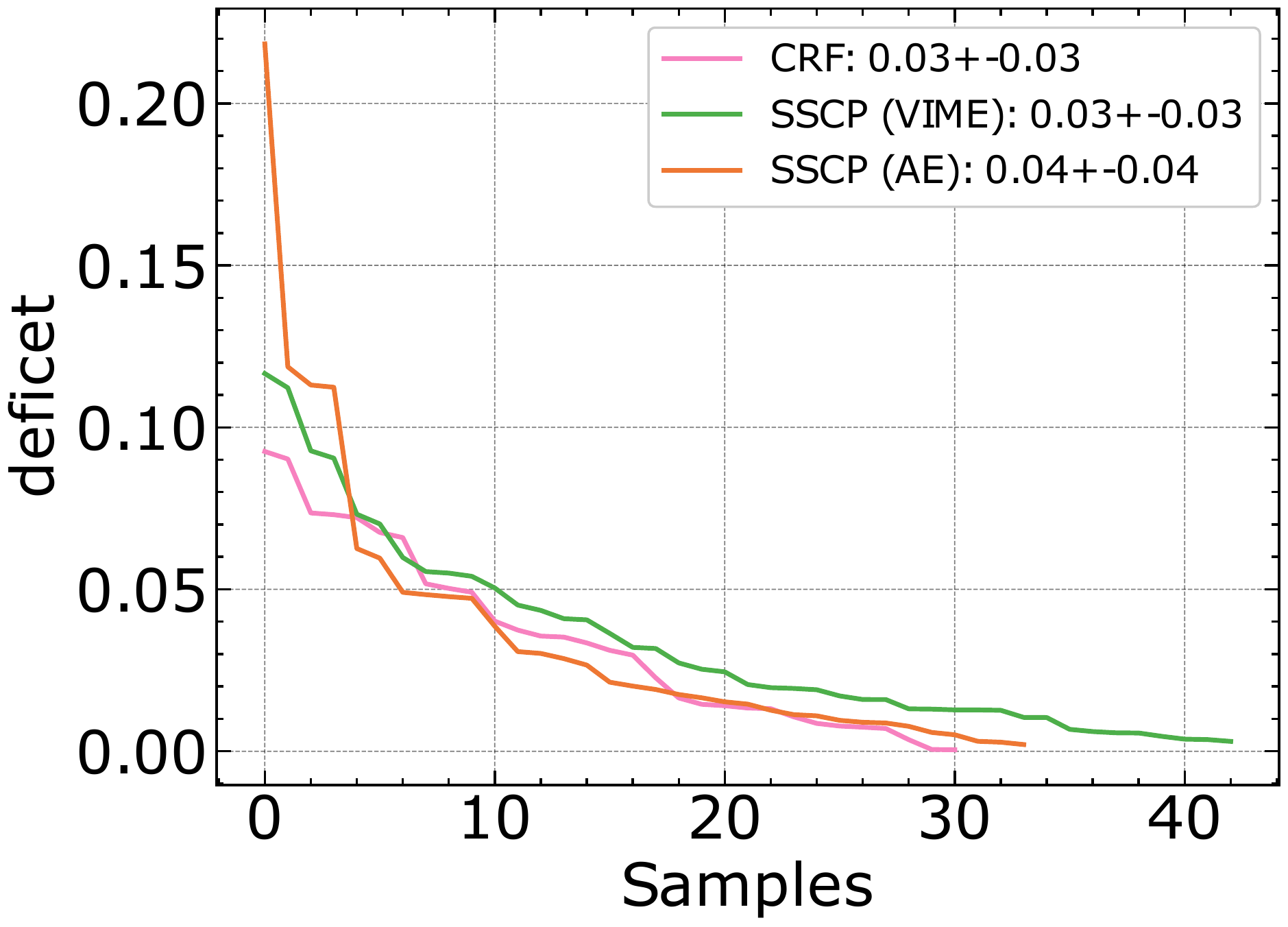}

        \caption{\footnotesize{Deficit}}
    \end{subfigure}%

    \caption{Comparing the metrics - ordered from the largest to smallest. We note that the benefit of SSCP for both VIME and AE is largely in reducing the excess width of the intervals on those samples for which it is greatest (i.e. the most inefficient intervals).}\label{fig:ordered_plots}
    \rule{\textwidth}{.5pt}
\end{figure*}

We then continue the deep-dive to better understand the performance improvements. We specifically analyze the types of samples for which the self-supervised loss provides the greatest gains. Figure \ref{fig:ordered_plots} orders the metrics sample-wise from largest to smallest. We can clearly see that the SSCP improvements are, in general, for those samples for which the metric is the highest in CRF. In other words, the self-supervision helps to improve the prediction intervals where the greatest inefficiency occurs on the most challenging samples with the greatest uncertainty. This is of course also the most natural target, as it is more challenging to improve prediction intervals that are already close to optimal.

\begin{figure*}[!h]

\centering
    \begin{subfigure}[t]{.33\textwidth}
        \includegraphics[width=\textwidth]{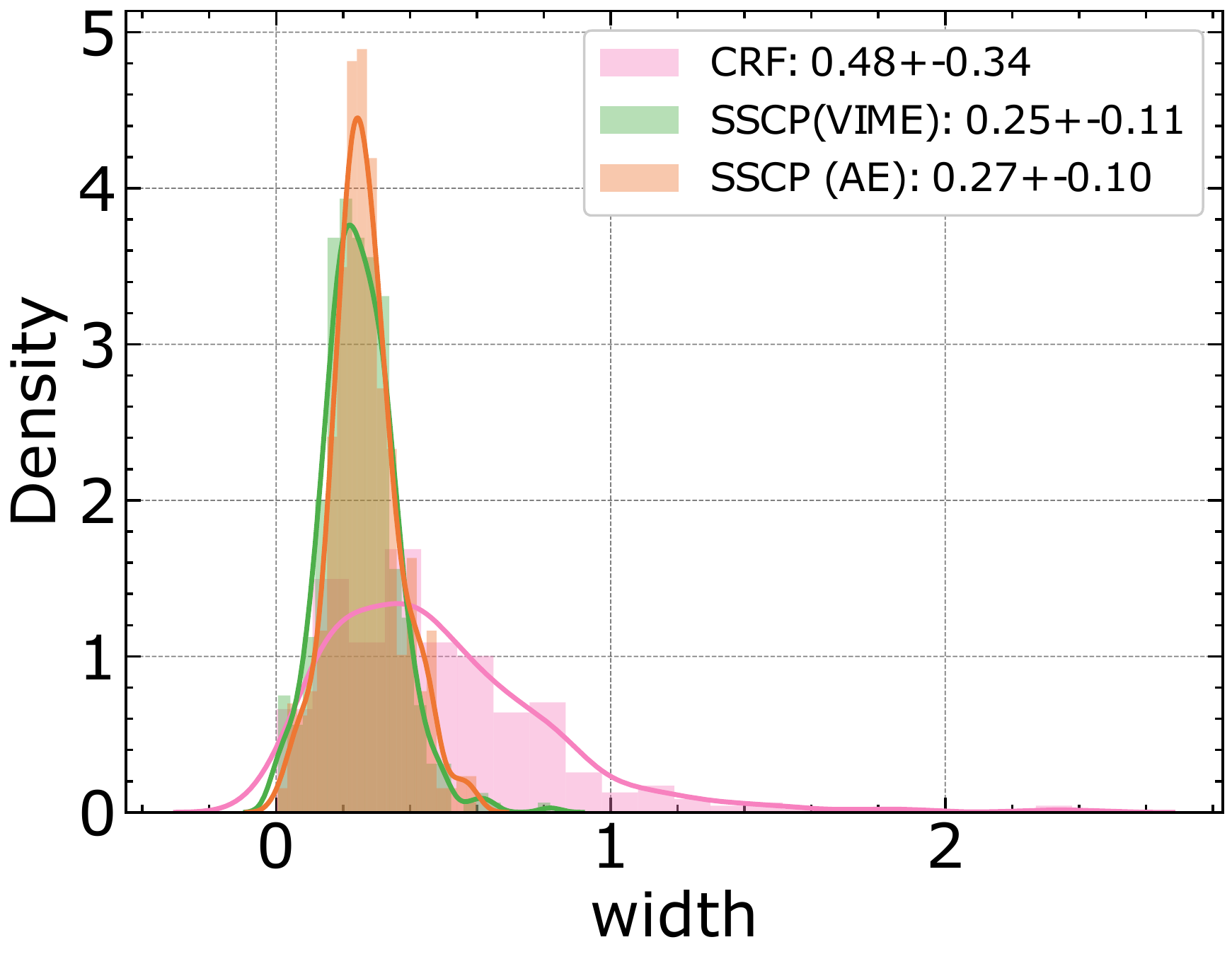}

        \caption{\footnotesize{Width.}}
    \end{subfigure}%
    ~
    \begin{subfigure}[t]{.33\textwidth}
        \includegraphics[width=\textwidth]{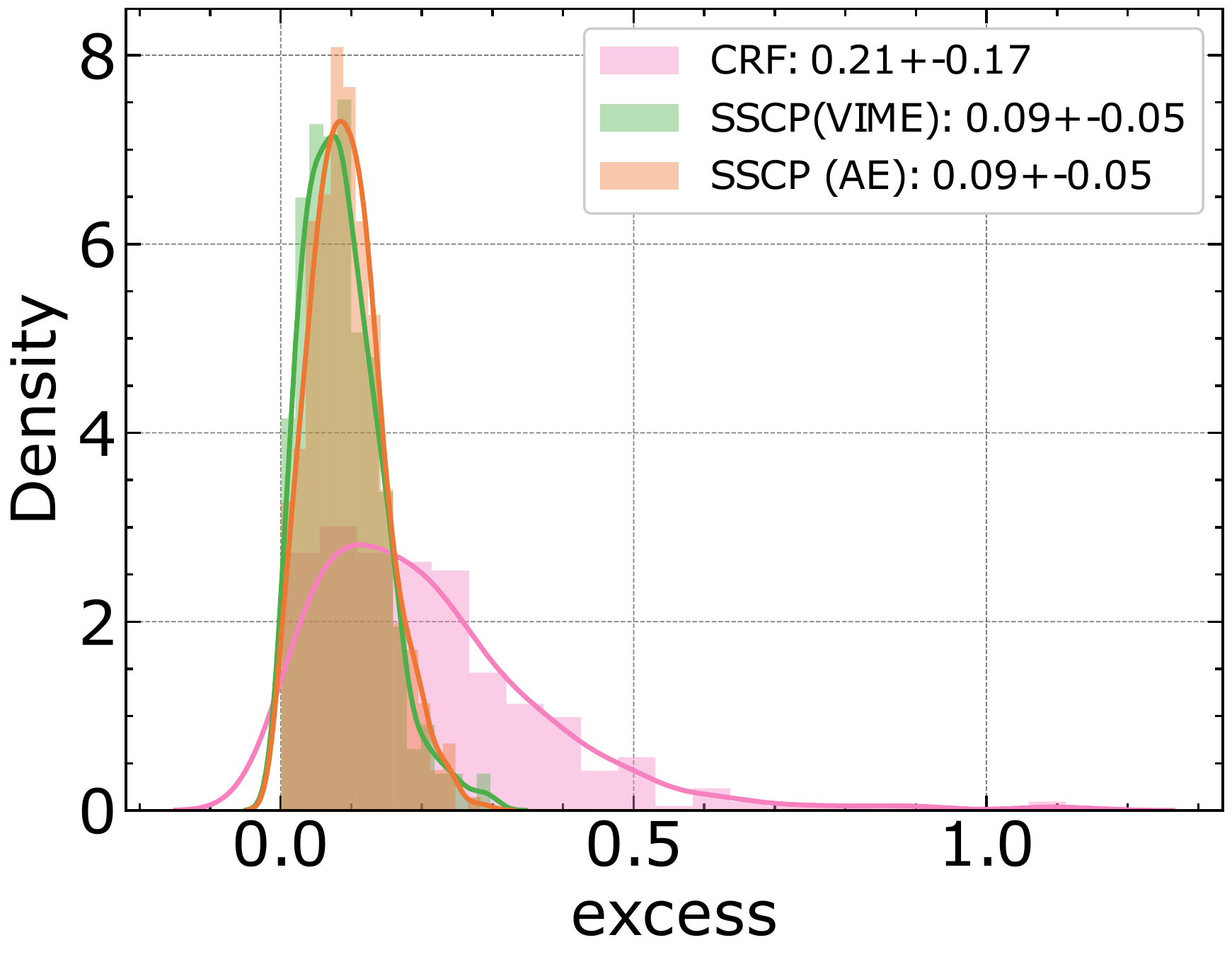}

        \caption{\footnotesize{Excess}}
    \end{subfigure}%
    ~
    \begin{subfigure}[t]{.33\textwidth}
        \includegraphics[width=\textwidth]{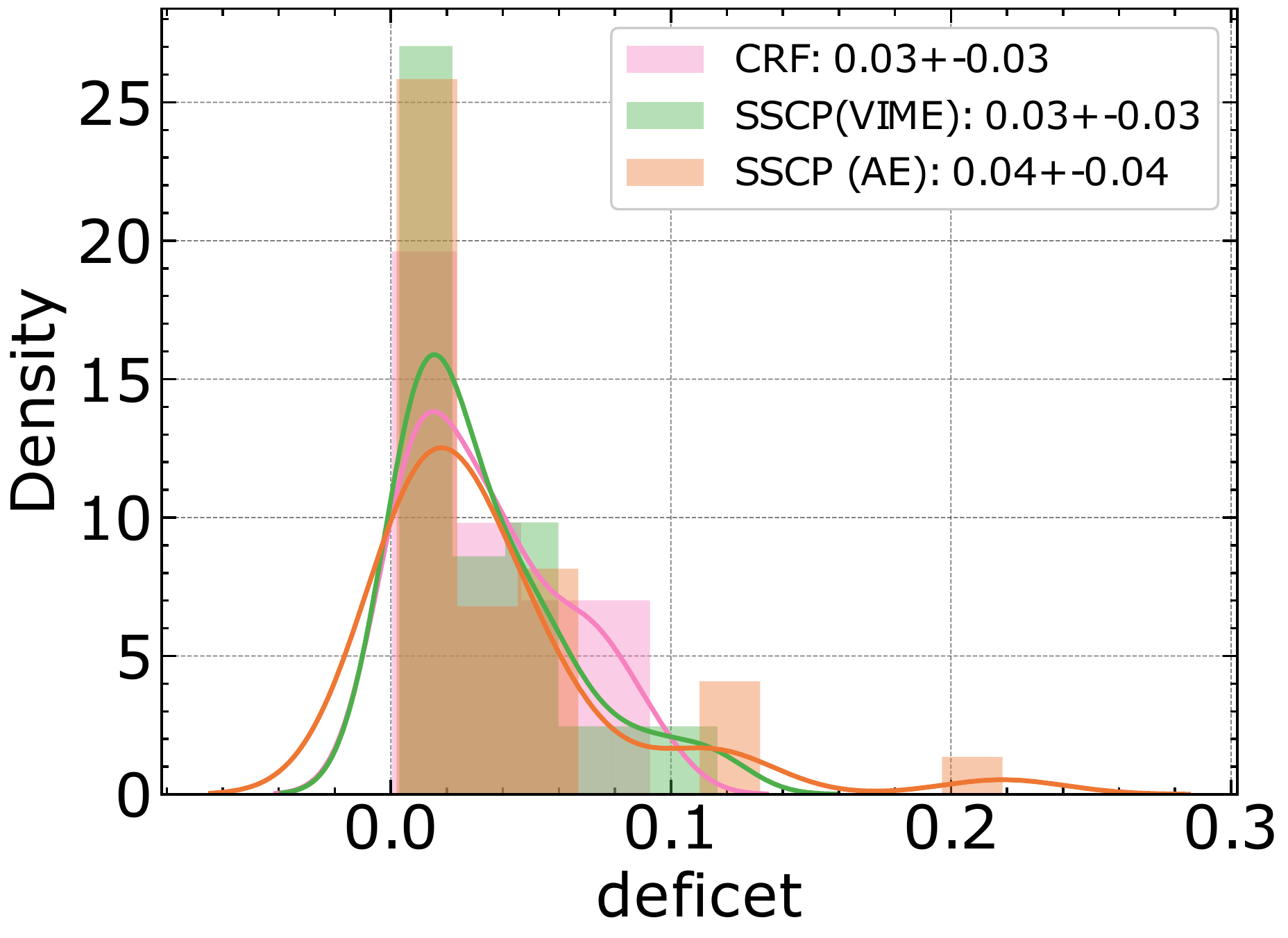}

        \caption{\footnotesize{Deficit}}
    \end{subfigure}%

    \caption{Comparison of the distributions of the metrics. We can clearly see that the addition of the self-supervision helps to reduce the long tails of the distributions corresponding to the most inefficient intervals.}\label{fig:distplots}
    \rule{\textwidth}{.5pt}
\end{figure*}

Figure \ref{fig:distplots} which illustrates the distribution of said metrics also confirms this insight. We note the reduction in the long tails of the distributions when augmenting with the self-supervised loss.

To summarize, in general, SSCP largely improves the intervals on samples with higher uncertainty (i.e. larger intervals) which often correspond to regions of sparsity in the input space. These fine-grained improvements at the sample-level then translate into the dataset-level improvements to the prediction intervals reported in our experiments (Section \ref{experiments}).

\subsubsection{Robustness study} \label{robustness-study}

\paragraph{Goal.} We wish to assess the robustness of the intervals and their improvements over the different random runs. Consequently, we aim to assess the confidence interval (CI) gains (i.e. reductions) for SSCP (VIME) vs. CRF.

\begin{figure}[!h]
    \centering
    \includegraphics[width=0.35\textwidth]{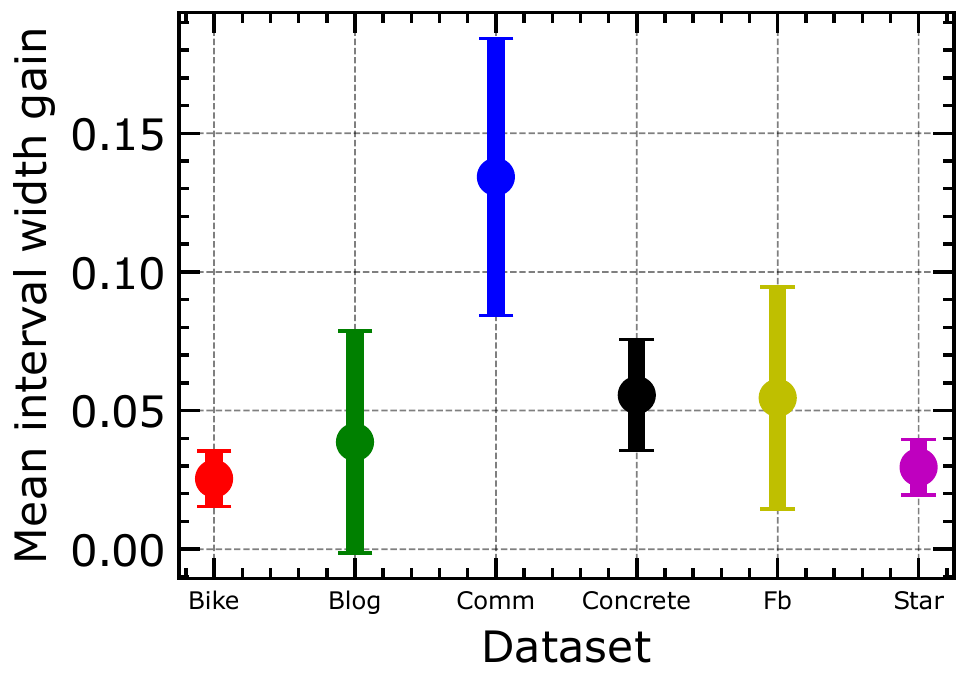}
    \caption{Assessing the 90\% confidence interval gains --- which implies width reductions across datasets, averaged across runs. We find that SSCP has a net reduction in intervals across all datasets, highlighting the robustness of the approach }
    \label{fig:intervals_gain}
\end{figure}

\paragraph{Analysis.}
We assess the 90\% CI's of the reduction in interval width (i.e. performance gain) across all datasets averaged across runs. The results are shown in Fig. \ref{fig:intervals_gain}.

We find that SSCP is robust in the improvements, showing a net gain over CRF (with interval reductions) across all datasets. This result is indicative of the robustness of including the self-supervised signal to improve the conformal intervals.

\end{document}

%% file: math_commands.tex
\usepackage{amsmath,amsfonts,bm}

\def\eqref#1{equation~\ref{#1}}

\def\1{\bm{1}}

\DeclareMathAlphabet{\mathsfit}{\encodingdefault}{\sfdefault}{m}{sl}
\SetMathAlphabet{\mathsfit}{bold}{\encodingdefault}{\sfdefault}{bx}{n}

\newcommand{\R}{\mathbb{R}}

